\documentclass[sigconf]{acmart}

\usepackage{enumitem}

\usepackage[english]{babel}

\usepackage{hyperref}       
\usepackage{url}            
\usepackage{booktabs}       
\usepackage{amsfonts}       
\usepackage{nicefrac}       
\usepackage{microtype}      
\usepackage{xcolor}         
\usepackage{amsmath}
\usepackage{algorithm}
\usepackage{algpseudocode}
\usepackage{subfiles}
\usepackage{graphicx}
\usepackage{tcolorbox}
\usepackage{textcomp,mathcomp}
\usepackage{float}
\usepackage{multirow}
\tcbuselibrary{breakable}
\tcbset{
    colback=blue!2!white, 
    colframe=black, 
    fonttitle=\bfseries,
    rounded corners,
    boxrule=1pt, 
    breakable, 
}

\usepackage{tikz}
\usepackage{subcaption}
\usetikzlibrary{trees}
\usetikzlibrary{mindmap, arrows.meta, shapes, arrows.meta, positioning}
\definecolor{darkblue}{HTML}{CDDCE5}
\definecolor{darkgreen}{HTML}{B9D5BA}
\definecolor{darkpink}{HTML}{F8DDDA} 
\definecolor{darkpurple}{HTML}{E1D7D0}

\usepackage{hyperref}
\hypersetup{
    colorlinks=true,
    linkcolor=blue,
    citecolor=blue,
    filecolor=blue,
    urlcolor=blue
}

\tcbuselibrary{skins}

\newcommand{\myhyperrefA}[2]{%
  \hyperref[#1]{%
    \tcbox[on line, colback=darkpurple, colframe=green, boxrule=0.5pt, boxsep=0pt, left=0.1pt, right=0.1pt, 
           top=0.2pt, bottom=0pt, sharp corners]{#2}%
  }%
}

\newcommand{\myhyperrefL}[2]{%
  \hyperref[#1]{%
    \tcbox[on line, colback=darkgreen, colframe=green, boxrule=0.5pt, boxsep=0pt, left=0.1pt, right=0.1pt, 
           top=0.2pt, bottom=0pt, sharp corners]{#2}%
  }%
}

\newcommand{\myhyperrefM}[2]{%
  \hyperref[#1]{%
    \tcbox[on line, colback=darkblue, colframe=green, boxrule=0.5pt, boxsep=0pt, left=0.1pt, right=0.1pt, 
           top=0.2pt, bottom=0pt, sharp corners]{#2}%
  }%
}

\newcommand{\myhyperrefS}[2]{%
  \hyperref[#1]{%
    \tcbox[on line, colback=darkpink, colframe=green, boxrule=0.5pt, boxsep=0pt, left=0.1pt, right=0.1pt, 
           top=0.2pt, bottom=0pt, sharp corners]{#2}%
  }%
}

\newcommand{\myhyperrefWhite}[2]{%
  \hyperref[#1]{%
    \tcbox[on line, colback=white, colframe=green, boxrule=0.5pt, boxsep=0pt, left=0.1pt, right=0.1pt, 
           top=0.2pt, bottom=0pt, sharp corners]{#2}%
  }%
}

\def\benchmark{\textbf{\textit{ChemEval}}}

\AtBeginDocument{%
  }

\setcopyright{acmlicensed}
\copyrightyear{2018}
\acmYear{2018}
\acmDOI{XXXXXXX.XXXXXXX}

\acmConference[Conference'24]{Make sure to enter the correct
  conference title from your rights confirmation emai}{2024}{}
\acmISBN{978-1-4503-XXXX-X/18/06}

\begin{document}

\title{\textit{ChemEval}: A Comprehensive Multi-Level Chemical Evalution for Large Language Models}


\author{%
    \begin{tabular}{c}
    \textbf{Yuqing Huang}$^{1, 2, }$\footnotemark[1], 
    \textbf{Rongyang Zhang}$^{1, 2, }$\footnotemark[1],
    \textbf{Xuesong He}$^{1, 2}$,
    \textbf{Xuyang zhi}$^{1, 2}$,
    \textbf{Hao Wang}$^{1, 2, }$\footnotemark[2],\\
    \textbf{Xin Li}$^{1, 2, 3, }$\footnotemark[2],
    \textbf{Feiyang Xu}$^{3}$,
    \textbf{Deguang Liu}$^{1}$,
    \textbf{Huadong Liang}$^{3}$,
    \textbf{Yi Li}$^{3}$,\\
    \textbf{Jian Cui}$^{3}$,
    \textbf{Zimu Liu}$^{3}$,
    \textbf{Shijin Wang}$^{2, 3, 4}$,
    \textbf{Guoping Hu}$^{2, 3}$,
    \textbf{Guiquan Liu}$^{1, 2, }$\footnotemark[2],\\
    \textbf{Qi Liu}$^{1, 2}$,
    \textbf{Defu Lian}$^{1, 2}$,
    \textbf{Enhong Chen}$^{1, 2, }$\footnotemark[2]
    \end{tabular}
}

\affiliation{%
    \vspace{1mm}
    \institution{$^1$University of Science and Technology of China, Hefei, China}
    \institution{$^2$State Key Laboratory of Cognitive Intelligence, Hefei, China}
    \institution{$^3$Artificial Intelligence Research Institute, iFLYTEK Co., Ltd., Hefei, China}
    \institution{$^4$iFLYTEK AI Research (Hebei), Langfang, China}
    \country{}
}

\email{{huangyuq, zhangry13}@mail.ustc.edu.cn, {wanghao3, leexin}@ustc.edu.cn, fyxu2@iflytek.com}

\renewcommand{\shortauthors}{Huang et al.}



\begin{abstract}


There is a growing interest in the role that LLMs play in chemistry which lead to an increased focus on the development of LLMs benchmarks tailored to chemical domains to assess the performance of LLMs across a spectrum of chemical tasks varying in type and complexity. However, existing benchmarks in this domain fail to adequately meet the specific requirements of chemical research professionals. To this end, we propose \textbf{\textit{ChemEval}}, which provides a comprehensive assessment of the capabilities of LLMs across a wide range of chemical domain tasks. Specifically, ChemEval identified 4 crucial progressive levels in chemistry, assessing 12 dimensions of LLMs across 42 distinct chemical tasks which are informed by open-source data and the data meticulously crafted by chemical experts, ensuring that the tasks have practical value and can effectively evaluate the capabilities of LLMs. In the experiment, we evaluate 12 mainstream LLMs on ChemEval under zero-shot and few-shot learning contexts, which included carefully selected demonstration examples and carefully designed prompts. The results show that while general LLMs like GPT-4 and Claude-3.5 excel in literature understanding and instruction following, they fall short in tasks demanding advanced chemical knowledge. Conversely, specialized LLMs exhibit enhanced chemical competencies, albeit with reduced literary comprehension. This suggests that LLMs have significant potential for enhancement when tackling sophisticated tasks in the field of chemistry. We believe our work will facilitate the exploration of their potential to drive progress in chemistry. Our benchmark and analysis will be available at {\color{blue} \url{https://github.com/USTC-StarTeam/ChemEval}}.

\end{abstract}






\maketitle

\renewcommand{\thefootnote}{\fnsymbol{footnote}} 
\footnotetext[1]{These authors contributed equally to this work.} 
\footnotetext[2]{Corresponding authors.} 









\section{Introduction}

The advent of large language models has ushered in a transformative era in artificial intelligence, particularly within the domain of natural language processing. The expansive capabilities of these models have not only redefined the boundaries of text generation and understanding\cite{brown2020language, ouyang2022training, touvron2023llama, achiam2023gpt} but have also opened new avenues for barious domains, such as recommendation\cite{wu2024survey, yin2024dataset, shen2024exploring, han2024efficient}, social\cite{wang2019mcne, wang2021hypersorec} and scientific exploration\cite{beltagy2019scibert, hong2022diminishing, bhattacharjee2024indus}. Researchers have adeptly employed LLMs to accelerate the pace of scientific research and instigate a transformative shift in scientific research paradigms. The field of chemistry has notably profited from the integration and advancement of LLMs\cite{yu2024llasmol, chen2024pharmgpt, zhang2021motif, hao2020asgn}, becoming a key area where these sophisticated technologies have delivered substantial advantages. The intricate nature of chemical research, involving complex molecular interactions and reactions, presents a unique challenge that LLMs are ready to address through advanced pattern recognition and predictive analytics.

In order to systematically assess the capabilities of LLMs across various domains and identify areas for their potential enhancement, numerous benchmarking initiatives have been introduced. For instance, the MMLU\cite{hendrycks2020measuring} covers 57 tasks spanning basic mathematics, American history, computer science, law, and other fields. The XieZhi\cite{gu2024xiezhi} benchmark includes three major academic categories with 516 specific subjects. However, general benchmarks\cite{zhong2023agieval, huang2024c} often overlook detailed assessment of chemical knowledge. Although Sun et al. introduce SciEVAL\cite{sun2024scieval} as a framework for assessing the competencies of LLMs within the scientific domain, the chemistry-related tasks are overly simplistic and do not adequately capture the depth required. Regarding chemistry domain-specialized benchmarks, Guo et al. \cite{guo2023can} propose eight chemical tasks aimed at assessing understanding, reasoning, and explanation abilities, but it consists of tasks derived from existing public datasets, which may be insufficient to capture the full spectrum of competencies needed for thorough chemical research. Other studies like \cite{white2023assessment, liu2023chatgpt} have similar problems. This limitation prevents them from tackling key issues of interest to chemistry researchers and has not fully met the specialized needs of chemistry. 

In light of these considerations, we introduce \benchmark, a benchmark designed to address the gap in the comprehensive assessment framework for LLMs in chemistry by providing a multi-dimensional evaluation. \textbf{1). Extensive} tasks are included in ChemEval, which encompasses chemical tasks of interest to researchers that were not included in previous benchmarks. It has four levels, twelve dimensions, and a total of forty-two distinct tasks, covering a vast array of issues within the domain of chemical research. \textbf{2). In-depth} tasks are also part of ChemEval, specifically designed to assess LLMs' capabilities in handling sophisticated chemical challenges. \textbf{3). Domain experts} in chemistry have meticulously crafted task datasets and prompts for ChemEval, partly addressing the previous lack of domain-specific data in chemistry benchmarks. Compared to previous work, our study encompasses a broader range of tasks that are of actual concern in chemical research. It assesses models on a graduated scale of capabilities, from general to domain-specific skills, to determine the model's proficiency. Our aim is to construct specialized tasks from the perspective of chemical researchers, thereby providing valuable insights for AI researchers and chemists, and improve large language models' effectiveness in chemical research.

For experiments, we conducted a highly detailed evaluation process, focusing on designing prompts that challenge LLMs, including system-specific prompts, task-specific prompts, and few-shot settings. We evaluated currently widely used LLMs, including both general LLMs and specialized chemical LLMs, and gained many meaningful insights. This comprehensive evaluation has revealed that though general LLMs like GPT-4\cite{achiam2023gpt} and Claude-3.5\cite{anthropic2024claude} excel in Literature Understanding tasks possess great instruction following capability, they struggle with tasks that require a deeper understanding of molecular structures and scientific inference. On the other side, specialized LLMs generally show improved chemical abilities even when their ability to understand literature and instruction following capability is diminished. This finding underscores the need for significant improvements in the way LLMs are trained and evaluated for chemical tasks.

We highlight the contributions of this paper as follows:
\begin{itemize}[leftmargin=*,align=left]
\item We have established an open-source benchmark, the first LLMs benchmark in chemistry that offers an evaluation of knowledge deduction, integrating extensive coverage with profound depth, fostering a collaborative environment for the scientific community to build upon our work and drive innovation in the application of LLMs to chemistry.

\item  We set up 4 progressive levels and access 12 model capability dimensions through 42 tasks in ChemEval, which is developed through extensive discussions and collaborative design with che-mistry researchers, involves constructing novel tasks of interest to chemical researchers and encompassing the primary focal points of chemical research.

\item We conducted a comprehensive evaluation of LLMs in chemical tasks, using various prompt settings to assess 12 LLMs, including both general and specialized LLMs. This revealed significant differences between general and specialized models and identified challenging tasks with potential for optimization. This work offers critical insights to guide researchers in the optimization and application of LLMs, thereby enhancing their effectiveness in chemical research.
\end{itemize}









\begin{table*}[t]
\caption{The statistics of \benchmark. It includes 4 progressive levels, evaluating 12 dimensions of LLMs capabilities and featuring 42 distinct chemical tasks that cover a wide range of chemical knowledge, from foundational concepts to advanced topics suitable for graduate-level research}
\vspace{-0.2cm}
\resizebox{\textwidth}{!} {
\begin{tabular}{l|l|l|l}
\toprule
\textbf{Levels} & \textbf{Dimensions} & \textbf{Tasks(Metric)} & \textbf{Types} \\ 
\hline
\multirow[l]{2}{*}{\begin{tabular}[l]{@{}l@{}}\myhyperrefWhite{3.1}{Sec 3.1} Advanced\\Knowledge\\Question\\Answering\end{tabular}} 
    & \myhyperrefWhite{3.1.1}{Sec 3.1.1} Objective Question  & Multiple Choice(Acc), Fill-in-the-Blank(Acc), True/False(Acc)  & \begin{tabular}[l]{@{}l@{}}Classification, \\Open-Ended Question\end{tabular} \\[8.8pt] \cline{2-4}
    & \begin{tabular}[l]{@{}l@{}}\myhyperrefWhite{3.1.2}{Sec 3.1.2} Subjective Question\\Answering\end{tabular}  & Short Answer(BLEU-2), Calculation(Acc)  &  \begin{tabular}[l]{@{}l@{}}Text Generation, \\Open-Ended Question\end{tabular}\\
\hline
\multirow[l]{2}{*}{\begin{tabular}[l]{@{}l@{}}\\ \myhyperrefWhite{3.2}{Sec 3.2} Literature\\Understanding\end{tabular}} 
    & \myhyperrefWhite{3.2.1}{Sec 3.2.1} Information Extraction  & \begin{tabular}[l]{@{}l@{}}Chemical Named Entity Recognition(F1), Chemical Entity Relationship Classification(F1),\\Synthetic Reaction Substrate Extraction(Acc), Synthetic Reaction Additive Extraction(F1),\\Synthetic Reaction Solvent Extraction(F1), Reaction Temperature Extraction(F1),\\Reaction Time Extraction(F1), Reaction Product Extraction(Acc),\\Characterization Method Extraction(F1), Catalysis Type Extraction(F1), Yield Extraction(F1)\end{tabular}   & Information Extraction \\ \cline{2-4}
    & \myhyperrefWhite{3.2.2}{Sec 3.2.2} Inductive Generation  & \begin{tabular}[l]{@{}l@{}}Chemical Paper Abstract Generation(BLEU-2), Research Outline Generation(BLEU-2),\\Chemical Literature Topic Classification(Acc), Reaction Type Recognition and Induction(F1)\end{tabular}  & \begin{tabular}[l]{@{}l@{}}Text Generation, \\Classification,\\ Information Extraction\end{tabular} \\
\hline
\multirow[l]{4}{*}{\begin{tabular}[l]{@{}l@{}}\\ \\ \myhyperrefWhite{3.3}{Sec 3.3} Molecular\\Understanding\end{tabular}} 
    & \begin{tabular}[l]{@{}l@{}}\myhyperrefWhite{3.3.1}{Sec 3.3.1} Molecular Name\\Generation\end{tabular}  & Molecular Name Generation from Text Description(BLEU)  & Text Generation \\ \cline{2-4}
    & \begin{tabular}[l]{@{}l@{}}\myhyperrefWhite{3.3.2}{Sec 3.3.2} Molecular Name\\Translation\end{tabular}  & \begin{tabular}[l]{@{}l@{}}IUPAC to Molecular Formula(Exact match), SMILES to Molecular Formula(Exact match),\\IUPAC to SMILES(Exact Match), SMILES to IUPAC(Exact Match),\\SMILES to SELFIES and SELFIES to SMILES Translation(Exact Match)\end{tabular}  & Name Translation \\ \cline{2-4}
    & \begin{tabular}[l]{@{}l@{}}\myhyperrefWhite{3.3.3}{Sec 3.3.3} Molecular Property\\Prediction\end{tabular}  & \begin{tabular}[l]{@{}l@{}}Molecule Property Classification(Acc), \\Molecule Property Regression(Rank)\end{tabular} & \begin{tabular}[l]{@{}l@{}}Classification, \\Regression Ranking\end{tabular} \\ \cline{2-4}
    & \myhyperrefWhite{3.3.4}{Sec 3.3.4} Molecular Description  & Physicochemical Property Prediction from Molecular Structure(BLEU-2)  & Text Generation \\
\hline
\multirow[l]{4}{*}{\begin{tabular}[l]{@{}l@{}}\\ \\ \\ \myhyperrefWhite{3.4}{Sec 3.4} Scientific\\Knowledge\\Deduction\end{tabular}} 
    & \begin{tabular}[l]{@{}l@{}}\myhyperrefWhite{3.4.1}{Sec 3.4.1} Retrosynthetic Analysis\end{tabular}  & \begin{tabular}[l]{@{}l@{}}Substrate Recommendation(F1), Synthetic Pathway Recommendation(Acc),\\Synthetic Difficulty Evaluation(RMSE)\end{tabular} & \begin{tabular}[l]{@{}l@{}}Regression Prediction, \\Reaction Prediction\end{tabular} \\ \cline{2-4}
    & \begin{tabular}[l]{@{}l@{}}\myhyperrefWhite{3.4.2}{Sec 3.4.2} Reaction Condition\\Recommendation\end{tabular}  & \begin{tabular}[l]{@{}l@{}}Ligand Recommendation(F1), Reagent Recommendation(F1),\\Solvent Recommendation(F1), Catalyst Recommendation(F1),\\Reaction Temperature Recommendation(RMSE), Reaction Time Recommendation(RMSE)\end{tabular}  & \begin{tabular}[l]{@{}l@{}}Regression Prediction,\\Reaction Prediction\end{tabular} \\ \cline{2-4}
    & \begin{tabular}[l]{@{}l@{}}\myhyperrefWhite{3.4.3}{Sec 3.4.3} Reaction Outcome\\Prediction\end{tabular} & \begin{tabular}[l]{@{}l@{}}Reaction Product Prediction(F1), Product Yield Prediction(Acc),\\Reaction Rate Prediction(Overlap)\end{tabular}  & \begin{tabular}[l]{@{}l@{}}Classification,\\Reaction Prediction,\\Range Overlap\end{tabular} \\ \cline{2-4}
    & \begin{tabular}[l]{@{}l@{}}\myhyperrefWhite{3.4.4}{Sec 3.4.4} Reaction Mechanism\\Analysis\end{tabular}  & Intermediate Derivation(Acc) & Open-Ended Question\\
\bottomrule
\end{tabular}
}
\vspace{-0.2cm}
\label{dataset}
\end{table*}

\setcounter{section}{1}

\section{Related Work}

\subsection{Large Language Models}
The advent of LLMs has marked a significant milestone in the field of Natural Language Processing (NLP). Over the past few years, there has been a surge in the development of proprietary models such as GPT-4\cite{achiam2023gpt} and Claude-3.5\cite{anthropic2024claude}, which have demonstrated remarkable capabilities in various NLP tasks. These models, through their extensive training on diverse datasets, have achieved unprecedented levels of performance, often outperforming human benchmarks in tasks like language translation, summarization, and quest-ion-answering. Concurrently, open-source models like LlaMA\cite{touvron2023llama} and ChatGLM\cite{du2021glm} series have emerged as viable alternatives, providing the research community with accessible tools to explore and innovate within the NLP domain. The success of these models is attributed to their massive scale, the vast amount of diverse data they have been trained on, and the architectural advancements that make them able to capture complex linguistic patterns and generate highly human-like text.

\subsection{Large Language Models for Chemistry}
As the ability of general LLMs has gained widespread recognition, researchers have endeavored to harness their power to assist in scientific research tasks. However, the application of these models in specialized domains such as chemistry meets with some challenges. The lack of domain-specific knowledge often leads to inadequate performance, particularly when dealing with tasks that involve technical jargon and numerical calculations. To address this gap, several approaches have been proposed. For instance, Galactica\cite{taylor2022galactica} was developed through extensive pre-training on scientific datasets, while SciGLM\cite{zhang2024sciglm} employed fine-tuning techniques using relevant datasets to enhance its performance in scientific tasks and ChemCrow\cite{bran2023chemcrow} augmented the LLM performance in chemistry by integrating 18 expert-designed tools. In the chemical domain, models like ChemDFM\cite{zhao2024chemdfm}, LlaSMol\cite{yu2024llasmol}, and ChemLLM\cite{zhang2024chemllm} have been introduced, each with tailored training regimes to imbue the models with chemical knowledge. Additionally, Drugchat\cite{liang2023drugchat} and Drugassist\cite{ye2023drugassist} have been specifically trained to understand molecular structures and chemical properties. Despite these efforts, the comprehensive understanding of the chemical domain by LLMs remains an area ripe for further exploration and development.

\subsection{Large Language Model Evaluations}

The progress made in the field of LLMs is tightly linked to the establishment of robust evaluation frameworks. For general tasks, benchmarks such as MMLU\cite{hendrycks2020measuring} and GLUE\cite{wang2018glue} have become standard tools for assessing model capabilities. In the scientific domain, recent initiatives like SciEval\cite{sun2024scieval}, SceMQA\cite{liang2024scemqa}, and SciAssess\cite{cai2024sciassess} have been introduced to evaluate scientific reasoning and knowledge. In the domain of chemistry, however, there are few compressive benchmarks available. For instance, ChemLLMbench\cite{guo2023can}, focusing on chemical task evaluation, closely resembles our work. It assesses only eight task categories, missing the broader competencies across chemical disciplines. Additionally, ChemLLMbench relies on publicly available datasets without expert quality review. Therefore, the chemical domain has yet to see the development of a comprehensive and systematic benchmarking suite. The absence of such benchmarks hinders the advancement of LLMs in chemistry, as it limits the ability to accurately measure and compare the performance of models tailored to chemical tasks. This can be attributed to the fact that chemistry encompasses a rich tapestry of intricate conceptual knowledge as well as complex computational tasks, coupled with a scarcity of domain-specific data in the field of chemistry. So the establishment of a domain-specific benchmark is crucial for driving innovation, facilitating the development of more sophisticated models, and ultimately, enhancing the integration of LLMs in chemical research and applications. In this case, we introduce {\benchmark}, a comprehensive, detailed and novel benchmark to assess the capabilities of LLMs in the domain of chemistry.









\section{ChemEval}
While the evaluation of large language models has been extensively conducted across domains such as law\cite{niklaus2023lextreme, chalkidis2021lexglue, fei2023lawbench}, finance\cite{islam2023financebench, zhang2023fineval, xie2023pixiu}, healthcare\cite{zhu2023promptcblue, wang2023cmb}, and sciences\cite{wang2023scibench, sun2024scieval, liang2024scemqa}, the domain-specific assessment within the field of chemistry remains notably sparse. Thus, we introduce a refined benchmark named {\benchmark} specifically designed to evaluate the capabilities of LLMs within the chemical domain to fill the absence of a holistic benchmark that encompasses the diverse range of tasks within the chemical domain. As illustrated in table \ref{dataset}, it contains four levels in the field of chemistry, each of which includes several different chemical dimensions, ensuring a comprehensive evaluation of LLMs. This framework measures the models' ability to understand and infer chemical knowledge from a broad range of dimensions through a series of meticulously designed tasks.

The following part provides a brief introduction to the four levels, with detailed explanations to be presented later.

\vspace{+0.2cm}

\noindent\textbf{1). Advanced Knowledge Question Answering: }
ChemEval initiates with the Advanced Knowledge Question Answering segment, which serves as the foundational layer. This level is meticulously designed to evaluate the LLM's understanding of core chemical concepts and principles, laying the groundwork for assessing the model's ability to comprehend and apply fundamental chemical knowledge accurately and efficiently.

\vspace{+0.2cm}

\noindent\textbf{2). Literature Understanding:}
Progressing from the basics, the Literature Understanding level evaluates the LLM's capacity to interpret and synthesize information from scientific literature. This segment demands the model to not only comprehend complex texts but also to extract, summarize, and critically analyze the content, reflecting its aptitude for learning from scholarly works. 

\vspace{+0.2cm}

\noindent\textbf{3). Molecular Understanding:}
The Molecular Understanding category advances ChemEval to the molecular level\cite{jablonka2024leveraging, hocky2024connecting, li2024empowering}, evaluating the LLM's analytical and computational capabilities concerning chemical entities. This task involves the interpretation of molecular structures, properties, and dynamics, requiring the mo-del to demonstrate a nuanced understanding of compounds and their interactions, which are crucial for advanced chemical research.

\vspace{+0.2cm}

\noindent\textbf{4). Scientific Knowledge Deduction:}
Culminating the ChemEval, the Scientific Knowledge Deduction level represents the summit of the evaluation, focusing on the model's innovative and evaluative capabilities in scientific research\cite{boiko2023autonomous}. This task challenges the LLM to generate hypotheses and synthesize new scientific insights for the purpose of scientific discovery.

In the following sections, we will provide a detailed introduction to the task content and data construction process of ChemEval. Except for tasks of Advanced Knowledge Question Answering are in Chinese, other tasks are all in English. The details of these tasks and all subsequent ones will be described in \myhyperrefWhite{appendixb}{Appendix B}.

\begin{figure*}[h!]
    \centering
    \includegraphics[width=1\linewidth]{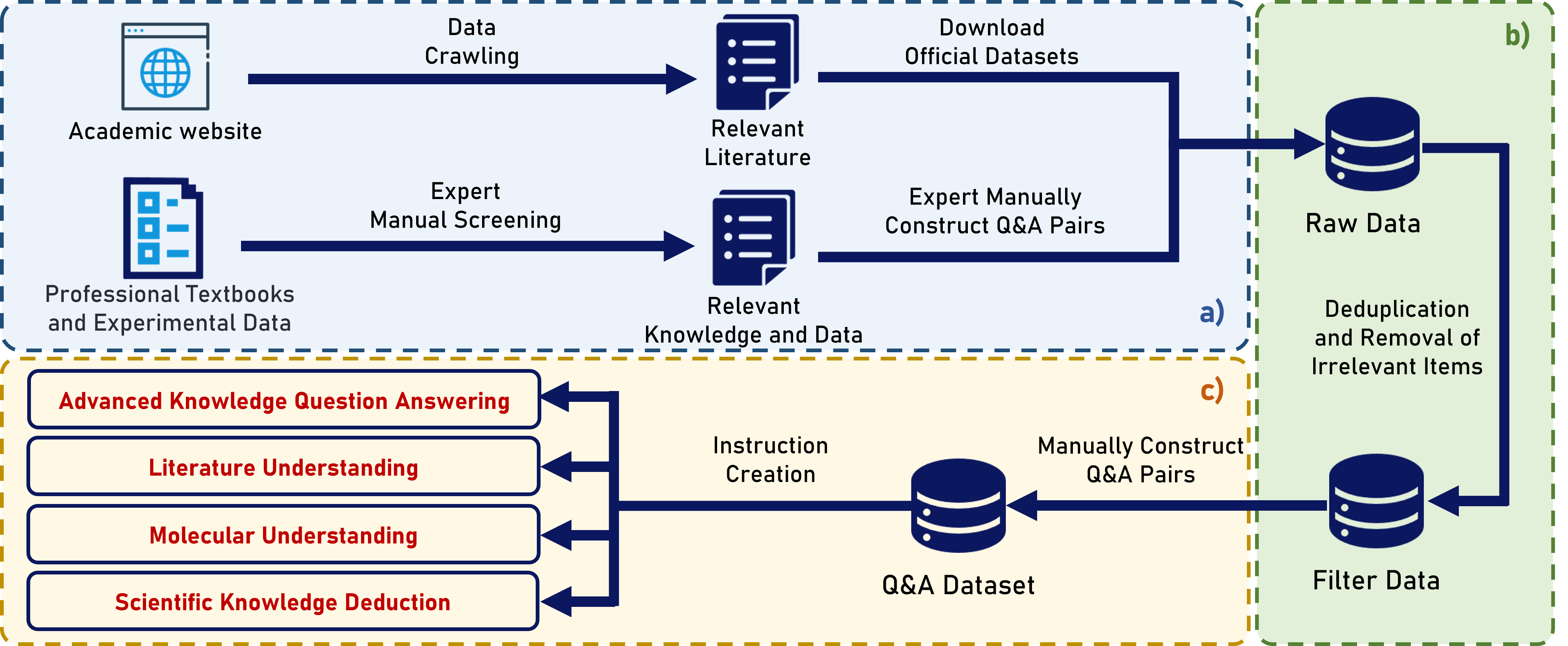}
    \caption{Data Collection steps of \benchmark. The process is divided into three main steps: a). Data Collection: Raw data is collected from academic websites via web crawling, and experts manually gather data from professional textbooks and experimental data. b). Data Filtering: The raw data undergoes deduplication and removal of irrelevant items to produce filtered data. c). Q\&A Pair Construction: Experts manually construct Q\&A pairs related to chemistry and create prompt instructions, resulting in four instruction test sets.}
    \label{fig 2}
\end{figure*}

\subsection{Advanced Knowledge Question Answering}
\label{3.1}
This segment is pivotal in assessing the models' proficiency in understanding and applying fundamental chemical concepts, which include \textit{Objective Question} dimension and \textit{Subjective Question} dimension, totally 5 different tasks. Through a blend of objective and subjective tasks, the Advanced Knowledge Question Answering challenges the models to demonstrate their insight in areas ranging from chemical terminology to quantitative analysis. The tasks within this section are designed to be both comprehensive and diagnostic, providing a clear measure of the models' readiness to tackle more advanced chemical inquiries. 

\subsubsection{\textbf{Objective Questions}}
\label{3.1.1}
The first dimension is objective question answering, which primarily assesses the model's grasp of fundamental chemical knowledge and its capability to apply this knowledge in straightforward scenarios. Objective question answering encompasses the following tasks: \textit{Multiple Choice Task}, \textit{Fill-in-the-Blank Task}, and \textit{True/False Task}. By incorporating these tasks, Chem-Eval can more effectively gauge the model's overall proficiency in understanding and applying chemical knowledge across various contexts and formats.

\subsubsection{\textbf{Subjective Questions}}
\label{3.1.2}
The second dimension is subjective question answering, which includes \textit{Short Answer Task} and \textit{Calculation Task}, both aiming to evaluate the depth of the model's comprehension and its ability to apply chemical knowledge effectively. Because on the basis of the previous task, the model also requires providing a detailed solution or reason, which involves the understanding of the chemical principles and concepts in the question, and applying these principles and concepts to construct logically clear and organized answers, which intuitively reflects the model's understanding of basic chemical knowledge. 

\subsection{Literature Understanding}
\label{3.2}
Advanced Knowledge Question Answering is designed to assess the model's comprehension and mastery of chemical knowledge, while Literature Understanding evaluates the model's capacity to interpret and assimilate information from chemical literature, which is foundational for subsequent inductive generation tasks. Literature Understanding, including \textit{Inductive Generation} dimension and \textit{Information Extraction} dimension, totally 15 tasks, delves into tasks crucial for understanding and extracting meaningful information from chemical literature. The primary focus is on assessing the LLMs' ability to accurately extract and interpret chemical data from texts, followed by generating new, contextually relevant content. The following subsections detail the specific tasks involved in this comprehensive evaluation.

\subsubsection{\textbf{Information Extraction}}
\label{3.2.1}
This is the first step to read a paper and also the foundation for the next inductive generation task. It involves the extraction of various elements related to chemistry, such as \textit{named entities}, \textit{reaction substrates}, and \textit{catalyst types}, encompassing a total of 11 tasks. These tasks aim to decompose and organize chemical information found in text, covering entities, relationships, and various aspects of chemical reactions.

\subsubsection{\textbf{Inductive Generation}}
\label{3.2.2}
Based on Information Extraction, Inductive Generation involves creating new, coherent, and contextually relevant content based on existing data and knowledge. This process incorporates \textit{Chemical Paper Abstract Generation}, \textit{Research Outline Generation}, \textit{Chemical Literature Topic Classification}, and \textit{Reaction Type Recognition and Induction}, all geared towards synthesizing and organizing chemical information in meaningful ways.

\subsection{Molecular Understanding}
\label{3.3}
This section builds upon the previous foundation to assess the mod-el's understanding and generative capabilities at the molecular level. It includes 4 dimensions: \textit{Molecular Name Generation}, \textit{Molecular Name Translation}, \textit{Molecular Property Prediction} and \textit{Molecular Description}, totally 9 tasks. Molecular Understanding explores tasks essential for molecular understanding, evaluating the LLMs' ability to generate, translate, and describe molecular names and properties. These tasks assess the models' proficiency in interpreting and generating chemical information accurately. The following subsections detail various specific tasks within this broader objective.

\subsubsection{\textbf{Molecular Name Generation}}
\label{3.3.1}
Molecular Name Generation is the basis of Molecular Understanding and only contains one task, \textit{Molecular Name Generation from Text Description}. This task is purposed to evaluate the capacity of LLMs in generating valid chemical structure representations. It necessitates that the models, based on intricate textual descriptions encompassing molecular structures, properties, and classifications, synthesize SMILES molecular formulas effectively.

\subsubsection{\textbf{Molecular Name Translation}}
\label{3.3.2}
Furthermore, Molecular Na-me Translation aims to enable a deep understanding of molecular structures and representations, which should serve as the fundamental knowledge for chemistry LLMs. It focuses on converting molecular names between different formats, requiring LLMs to output a specified alternative format based on a given molecular representation. It involves the conversion between representations of molecules such as \textit{IUPAC names} and \textit{SMILES}\cite{weininger1988smiles} molecular formulas, encompassing a total of five tasks, each focusing on distinct aspects of molecular notation conversion.

\subsubsection{\textbf{Molecular Property Prediction}} 
\label{3.3.3}
Apart from molecular na-me understanding, the ability to predict molecular property is also important. Molecular Property Prediction targets the forecast of a wide range of physical, chemical, and biological attributes of molecules, encapsulated in two core objectives: \textit{Molecule Property Classification}, which predicts categories of properties such as ClinTox, HIV inhibition, and polarity; and \textit{Molecule Property Regression}, focusing on estimating numerical values such as Lipophilicity, polarity and boiling point.

\subsubsection{\textbf{Molecular Description}} 
\label{3.3.4}
To further understand molecular, the Molecular Description task has been designed to evaluate LLMs' capability in understanding and describing molecular structures. This task consists of a single subtask: \textit{Physicochemical Property Prediction from Molecular Structure}.

\subsection{Scientific Knowledge Deduction}
\label{3.4}
Having established a solid grasp of basic chemical knowledge, the skill to interpret scientific literature, and the capacity to understand molecular structures, we expect that the model will proceed to conduct deeper chemical reasoning and deduction. So the part of Scientific Knowledge Deduction encompasses four key dimensions: \textit{Retrosynthetic Analysis},  \textit{Reaction Condition Recommendation}, \textit{Reaction Outcome Prediction} and \textit{Reaction Mechanism Analysis}, totally 13 tasks, which are essential for effective chemical synthesis. This part evaluates the LLMs' capabilities in retrosynthetic analysis, recommending reaction conditions, predicting reaction outcomes, and analyzing reaction mechanisms. These tasks provide a comprehensive assessment of the models' performance in these critical areas of chemical synthesis.

\subsubsection{\textbf{Retrosynthetic Analysis}}
\label{3.4.1}
Retrosynthetic Analysis is a crucial technique in the field of chemical synthesis, particularly in organic synthesis. It starts from the target product and analyzes possible synthesis pathways and reactant substrates, demonstrating the reverse reasoning ability of the large model in the field of chemical synthesis. It comprises \textit{Substrate Recommendation}, \textit{Synthetic Pathway Recommendation} and \textit{Synthetic Difficulty Evaluation}.

\subsubsection{\textbf{Reaction Condition Recommendation}} 
\label{3.4.2}
Based on the results of Retrosynthetic Analysis, LLMs can recommend suitable reaction conditions. Reaction condition recommendation is a key task in chemical synthesis, involving selecting the most suitable conditions for specific chemical reactions to ensure maximum efficiency, selectivity, and yield. This task integrates recommendations for conditions such as \textit{ligands}, \textit{reagents}, and \textit{catalysts}, encompassing a total of six tasks, each targeting a specific component of the reaction condition optimization.

\subsubsection{\textbf{Reaction Outcome Prediction}}
\label{3.4.3}
After determining the reaction pathway and reaction conditions, the large model can predict possible reaction outcomes. Reaction outcome prediction is a core technology in chemical synthesis aimed at predicting possible results of a reaction before it is actually carried out. This encompasses \textit{Reaction Product Prediction}, \textit{Product Yield Prediction}, \textit{Reaction Rate Prediction}.

\subsubsection{\textbf{Reaction Mechanism Analysis}} 
\label{3.4.4}
Reaction Mechanism Analysis is a critical area in the study of chemical reactions, aiming to explain the detailed steps involved in the transformation from reactants to products. This is the final step in the field of chemical synthesis, including identifying various intermediates, transition states, as well as the kinetic and thermodynamic parameters of each step in the reaction. \textit{Intermediate Derivation} is the sole subtask in this phase.

\begin{table*}[h!]
\Large
\caption{Comparative Analysis of 0-Shot Performance Among Mainstream LLMs Across 4 Levels of Chemical Evaluation}
\resizebox{\textwidth}{!}{
\begin{tabular}{ccccccccccccc}
\toprule
Metric & \textbf{GPT-4} & \textbf{\begin{tabular}[c]{@{}c@{}}claude-3.5-\\ Sonnet\end{tabular}} & \textbf{ERNIE-4.0} & \textbf{Kimi} & \textbf{\begin{tabular}[c]{@{}c@{}}LLaMA3- \\ 8B\end{tabular}} & \textbf{\begin{tabular}[c]{@{}c@{}}LLaMA3-\\  70B\end{tabular}} & \textbf{GLM-4} & \textbf{\begin{tabular}[c]{@{}c@{}}DeepSeek-\\ V2\end{tabular}} & \textbf{ChemDFM} & \textbf{\begin{tabular}[c]{@{}c@{}}LlaSMol-\\ Mistral\end{tabular}} & \textbf{\begin{tabular}[c]{@{}c@{}}ChemLLM-\\ 7B\end{tabular}} & \textbf{\begin{tabular}[c]{@{}c@{}}ChemSpark\end{tabular}} \\ \midrule
\multicolumn{13}{c}{\textit{\textbf{Advanced Knowledge Question Answering}}} \\ \midrule
Accuracy & 51.25 & \textbf{56.25} & 53.75 & 42.50 & 38.75 & 45.00 & 42.50 & 53.75 & 31.25 & 17.50 & 3.75 & 41.25 \\
BLEU-2 & 10.53 & 13.71 & 14.81 & 9.41 & 5.12 & 6.01 & 12.01 & \textbf{45.15} & 12.24 & 6.31 & 0.15 & 18.28 \\ \midrule
\multicolumn{13}{c}{\textit{\textbf{Literature Understanding}}} \\ \midrule
F1 & 54.71 & 60.35 & \textbf{60.62} & 51.56 & 58.36 & 60.00 & 50.64 & 63.87 & 37.66 & 2.28 & - & 43.90 \\
Accuracy & 34.12 & 22.44 & 19.89 & 55.33 & 30.32 & 59.26 & \textbf{64.66} & 43.46 & 28.66 & 8.33 & 11.67 & 22.95 \\
BLEU-2 & 33.56 & 33.00 & - & \textbf{63.64} & 33.66 & 31.22 & 35.93 & 32.80 & 38.20 & 0.55 & 0 & 3.31 \\ \midrule
\multicolumn{13}{c}{\textbf{Molecular Understanding}} \\ \midrule
BLEU & 40.21 & 40.80 & 44.99 & 8.86 & 13.85 & 43.11 & 23.89 & 24.43 & 75.91 & 75.65 & 0.94 & \textbf{80.00} \\
Exact Match & 1.00 & 10.00 & 11.00 & 2.00 & 0 & 3.00 & 3.00 & 0 & 16.00 & 14.00 & 0 & \textbf{71.00} \\
Accuracy & 56.50 & 53.60 & 67.65 & 55.05 & 51.60 & 44.75 & 48.10 & 55.55 & 60.95 & 39.00 & 26.00 & \textbf{76.75} \\
Rank & 6.00 & \textbf{2.86} & 5.43 & 8.57 & 6.14 & 4.29 & 6.86 & 7.71 & 8.57 & 8.14 & 9.71 & 4.00 \\
BLEU-2 & 13.87 & 19.74 & 28.98 & 47.28 & 35.54 & 39.64 & 41.97 & 40.86 & 38.86 & \textbf{52.78} & 0 & 50.44 \\ \midrule
\multicolumn{13}{c}{\textit{\textbf{Scientific Knowledge Deduction}}} \\ \midrule
F1 & 17.06 & 7.60 & 18.41 & 16.54 & 7.94 & 9.44 & 15.63 & 15.43 & 9.28 & 5.83 & - & \textbf{36.13} \\
Accuracy & 39.17 & 29.71 & 27.50 & \textbf{44.17} & 21.67 & 35.83 & 22.50 & 31.67 & 18.33 & 9.17 & 16.67 & 34.17 \\
RMSE(Valid Num) & 91.30(40) & 20.08(40) & 20.53(38) & 24.83(35) & 21.96(40) & 17.00(60) & 26.08(40) & 132.22(35) & 44.65(40) & 36.26(47) & 75.90(48) & \textbf{9.41(59)} \\
Overlap & 14.54 & \textbf{23.50} & 11.25 & 17.06 & 8.42 & 5.97 & 6.70 & 10.77 & 4.71 & 0 & 5.44 & 4.21 \\ \bottomrule
\end{tabular}
}
\vspace{-0.2cm}
\label{tab:0shot}
\end{table*}

\subsection{Evaluation}
\subsubsection{\textbf{Data Collection}}
Data plays an indispensable role in the realm of LLMs\cite{yin2024entropy}. Our data collection is comprised of two components: Open-source Data and Domain-Experts data. Open-source Data is based on keywords such as chemistry, large models, knowledge question answering, and information extraction, retrieve and download relevant papers on chemical large models from academic websites. Then, extract and code the downstream tasks and their datasets within the chemical evaluation system from the papers\cite{edwards2022translation, chen2023g, zhou2023uni}. Next, download the official datasets for the different downstream tasks, using the presence of an official test set as the main criterion for selection. Nevertheless, the scope of open-source data is inadequate, which is why we collect expert datasets to enhance the evaluation's rigor and breadth.
Domain-experts data is from scientific literature in the field, professional textbooks and supplementary materials, and laboratory chemical experiment data, manually construct question-answer pairs according to the task type.

\vspace{-0.1cm}

\subsubsection{\textbf{Data Processing}}
Through our data collection endeavors, we get a vast array of raw data in the chemical domain. However, to harness this data for our benchmarking work, it necessitates a subsequent phase of meticulous selection and filtration aligned with the diverse tasks.

Our data processing for different levels: \textit{1). Advanced knowledge question-answering.} We meticulously compile question-answer pai-rs derived from undergraduate and postgraduate level textbooks, as well as ancillary educational materials. These pairs encompass a broad spectrum of seven distinct categories: organic chemistry, inorganic chemistry, materials chemistry, analytical chemistry, biochemistry, physical chemistry, and polymer chemistry. This comprehensive selection ensures a diverse representation of chemical concepts and principles.
\textit{2). Literature understanding component.} We extract relevant fragments and questions from scientific literature, combining them with task-specific answers to create question-answer test sets for various downstream tasks. 
\textit{3). Molecular understanding and scientific knowledge deduction.} Our approach leverages a combination of open datasets and proprietary laboratory data sourced from our collaborating universities. We engage in the thoughtful design and construction of test sets meticulously aligned with the unique content requirements of downstream tasks. 

It is important to highlight that when integrating multiple open-source datasets for downstream tasks, we adopt a methodical approach to constructing the corresponding test sets. This involves employing proportional sampling techniques that take into account the varying scales of the different data sources. This strategy ensures that the test sets accurately reflect the broader dataset while maintaining a balanced distribution of question and answer types.

\vspace{-0.1cm}

\subsubsection{\textbf{Data Statistics}}
For each downstream task, a test set of 20 question-answer pairs and a few-shot set of 3 task introduction examples were constructed, all described in natural language text. Additionally, the molecular property classification task includes 100 items, while the molecular property regression task includes 140 items. Through our data collection endeavors, we get a vast array of raw data in the chemical domain. Notably, the test sets for different downstream tasks were cross-checked to remove duplicates with the training sets of corresponding tasks in open-source domain models, ensuring that there is no risk of data leakage in the evaluation of different downstream tasks.

\vspace{-0.1cm}

\subsubsection{\textbf{Instruction Creation}}
To evaluate the effectiveness of the model, in this paper, we constructed five sets of instruction sets for different downstream tasks: system-only instructions, task-specific prompts, and task-specific prompts with 1 to 3 example sets added respectively\cite{wei2022chain}. For downstream tasks with open-source datasets, to facilitate evaluation, the evaluation system in this paper strengthens the format of the output data based on its instructions. For the domain expert-built part, the evaluation system in this paper will design instructions for task introduction and formatted output according to the task type, and continuously adjust the instructions based on the return results of GPT-4, thereby strengthening the instructions for different self-constructed downstream tasks.

\vspace{+0.3cm}

\subsubsection{\textbf{Metrics}}
In this study, we utilize a range of evaluation metrics to comprehensively assess our models' performance across diverse tasks. For the majority of tasks, we utilize the F1 score and Accuracy. In addition, we utilize BLEU\cite{papineni2002bleu}, Exact Match, RMSE(Valid Num), Rank and Overlap in different tasks to meet the needs of different tasks. It is worth noting that Valid Num refers to the number of valid outputs by models and the value of RMSE is obtained through the weighted average of valid output. For some tasks with short answers, we only use 2-gram BLEU to evaluate the answers. For specific tasks like synthetic pathway recommendation, our evaluation combines automated metrics with expert manual review to ensure accuracy and professional insight. This framework ensures a detailed and effective evaluation of model performance across different settings. More detailed information about metrics 
as illustrated in table \ref{dataset}.

\benchmark, composed of the above series of tasks and each task builds upon the previous in a layered and progressive manner, gradually broadening the scope of chemical knowledge encompassed, increasing in difficulty, and deepening the comprehension of the intrinsic principles involved. Utilizing our ChemEval, model developers are equipped to discern and enhance the efficacy of their models, thereby facilitating targeted improvements and optimization. It also provides reliable technical support for research, education, and industrial applications, thereby promoting the advancement and innovation of LLMs for chemical knowledge inference. To achieve a more in-depth understanding of ChemEval, in the forthcoming section, we will evaluate the performance of prevalent LLMs on our ChemEval.








\begin{table}[t]
\Large
    \caption{Models evaluated in this paper.}
    \label{tab:Models}
    \resizebox{0.48\textwidth}{!}{
    \begin{tabular}{cccccc}
    \toprule
    \textbf{Model} & \textbf{Creator} & \textbf{Parameters}  & \textbf{Access} & \textbf{Chinese-oriented} & \textbf{Specialized}  \\
    \midrule
     GPT-4  & OpenAI & Undisclosed & API &  &   \\
     Claude-3.5-Sonnet & Anthropic & Undisclosed & API  &  &  \\
     ERNIE-4.0 & Baidu & Undisclosed & API  &$\checkmark$  &  \\
     Kimi &  Moonshot AI & Undisclosed & API  & $\checkmark$ &  \\
     GLM-4  &  ZhipuAI & Undisclosed & API  & $\checkmark$ &  \\
     DeepSeek-V2 & DeepSeek & 236B& API & $\checkmark$ &  \\
     LLaMA3-8B  & Meta & 8B & Weights  &  &  \\
     LLaMA3-70B & Meta & 70B & Weights  &  &  \\
     ChemDFM & OpenDFM &  13B & Weights  &  & $\checkmark$  \\
     LlaSMol & Osunlp & 7B & Weights & & $\checkmark$  \\
     ChemLLM & AI4CHem & 7B & Weights &  & $\checkmark$ \\
     ChemSpark & iFLYTEK & 13B &  weights   & $\checkmark$ & $\checkmark$  \\
    \bottomrule
    \end{tabular}
    }
    \vspace{-0.5cm}
\end{table}

\setcounter{section}{3}
\section{Experiment}
\subsection{Setup}

To comprehensively evaluate the chemical capabilities of LLMs, our evaluation framework includes assessments of most of the current general large models, as well as some recently fine-tuned models with a focus on chemical knowledge. As a representative of the general large model, GPT-4\cite{achiam2023gpt} is the best model from OpenAI that has undergone pretraining, instruction-tuning, and reinforcement learning. Claude-3.5, developed by Anthropic, is the latest iteration of the Claude model family and is often regarded as surpassing GPT in terms of performance. Claude-3.5-Sonnet\cite{anthropic2024claude}, the first release in this series, sets a new industry standard for intelligence. Baidu's ERNIE\cite{sun2021ernie} offers significant advancements in AI-driven content creation, while Kimi\cite{qin2024mooncake} by Moonshot AI can provide accurate responses in both English and Chinese. Meta AI's LLaMA\cite{touvron2023llama} is probably the best open-weight foundation model so far. We evaluate LLaMA3-8B and LLaMA3-70B here. GLM-4\cite{glm2024chatglm} by ZhipuAI outperforms LLaMA3-8B in various evaluations, and DeepSeek-V2\cite{deepseekv2} by DeepSeek is a robust Mixture-of-Experts (MoE) open-source Chinese language model comparable to GPT-4-turbo.

In the field of chemistry, specialized LLMs have demonstrated significant advancements. ChemDFM\cite{zhao2024chemdfm}, based on LLaMA-13B, can surpass GPT-4 on a great portion of chemical tasks, despite the significant size difference. LlaSMol\cite{yu2024llasmol} advances LLMs for chemistry through instruction fine-tuning of pre-trained models, with Mistral being the best base model. LlaSMol significantly outperforms Claude-3.5-Sonnet on most chemical tasks. ChemLLM\cite{zhang2024chemllm} by AI4Chem interprets and predicts chemical properties and reactions based on molecular structures, effectively analyzing complex chemical data to provide insights into molecular behavior and interactions. ChemSpark is trained through full-parameter fine-tunin-g based on the Spark 13B\footnote{https://www.xfyun.cn} foundation model by the dataset mixed of general domain Q\&A and chemical domain-specific Q\&A.

To illustrate the capability of LLMs in solving various chemical tasks, we present the average performance of LLMs across four levels under zero-shot condition, along with a detailed account of the zero-shot results. Due to the constraints on space, the average result of zero-shot results of all the models are shown in table \ref{tab:0shot} and detailed results of all subtasks representing different types of tasks are shown in \myhyperrefWhite{Appendix A.1}{Appendix A.1}. Additionally, to investigate the adaptability and in-context learning abilities of LLMs for chemical tasks, we report the average performance across the same four levels under three-shot conditions. We provide the detailed result of few-shot setting in \myhyperrefWhite{Appendix A.2}{Appendix A.2}.

\subsection{Performance Results}

We evaluate the model's performance by averaging the metrics for each task across four assessment dimensions. Certain models are unable to address specific tasks entirely. For example, ChemLLM demonstrates particularly poor instruction-following capabilities, which significantly impairs its ability to generate responses based on task prompts. Consequently, we are unable to provide numerical results for the tasks affected by this limitation. We discuss the key findings from our benchmark and analyze them to explore how different settings related to LLMs affect performance and provide valuable insights into Chemical benchmarks.

\subsubsection{\textbf{The models' performance across four levels.}}

Observing the models' performance across four levels, we have the following findings: \textit{1). Basic Knowledge:} The results indicate that general large models like DeepSeek-V2 and LLaMA3-70B excel in advanced knowledge answering, chemical literature comprehension, and scientific knowledge deduction tasks due to their extensive pre-training and document comprehension capabilities. DeepSeek-V2, in particular, shows strong performance across various tasks, while models like GPT-4 and Claude-3.5-Sonnet also perform well in certain areas. However, models like LlaSMol and ChemLLM struggled, highlighting challenges in instruction fine-tuning. \textit{2). Chemical expertise:} Chemistry-specific models like ChemSpark stand out in tasks requiring deep chemical knowledge, such as molecular understanding and scientific knowledge deduction, where they outperform general-purpose models, emphasizing the importance of specialized training for these tasks. Most models perform poorly on molecular name translation tasks, with performance metrics approaching zero. A key reason for this deficiency is that LLMs lack strict formatting constraints in their outputs, making it difficult to generate accurate molecular formulas consistently. In contrast, ChemSpark's training data includes a wide range of chemical literature and papers with various molecular formula formats, thus enhancing its adaptability and performance on this task.

\subsubsection{\textbf{The benefits and drawbacks of specialized-LLMs.}}

Compared with general-LLMs, we notice that the specialized-LLMs perform differently. \textit{1). The drawbacks of specialized-LLMs:} In advanced knowledge answering and literature comprehension tasks, chemical models perform significantly worse than general models. Although specialized models acquire domain-specific knowledge thr-ough fine-tuning, their foundational natural language processing capabilities are compromised. This suggests that these models may encounter challenges related to catastrophic forgetting during the fine-tuning process. \textit{2). The benefits of specialized-LLMs:} In specific tasks that require specialized terminology and molecular properties, chemical models tend to have an advantage. The limited proportion of specialized chemical data in the pre-training datasets of general models allows them to perform adequately on simpler tasks. However, when faced with more complex scenarios, their ability to process and infer specialized chemical knowledge is notably deficient. \textit{3). Instruction-following ability:} During the evaluation process, different models exhibited varying levels of instruction-following capability. The instruction-following ability of chemistry-specific LLMs was significantly lower than that of general LLMs. These models, while possessing deep knowledge in chemistry, may not have been as widely exposed to the variety of tasks and diverse data present in the benchmark, leading to difficulties in adapting to new or varied instructions.

\subsubsection{\textbf{The influence of few-shot.}} 

We find that few-shot prompting has a great impact on the model. \textit{1). Text processing capability:} Comparing results among 0-shot and few-shot, GPT-4 and ERNIE-4.0 get great performance enhancement in objective question answering while other models almost remain the same. 
When it comes to chemical literature comprehension and objective question answering, we find that few-shot prompting helps many models achi-eve better results, which indicates that few-shot helps improve the model's text comprehension, processing, and generalization abilities. But few-shot setting hurts the performance of ChemDFM, LlaSMol and ChemLLM. This is maybe because these models have not appropriately incorporated few-shot demonstrations into the instruction tuning stage, thus sacrificing few-shot in-context learning performance to obtain enhanced zero-shot instruction-following abilities. \textit{2). Reasoning ability:} As for Molecular Understanding and Scientific Knowledge Deduction, the improvement of all models is not particularly significant, and many indicators even have a decrease. It may stem from the intrinsic complexities of the tasks, potential mismatches between training data and task requirements, inadequate fine-tuning processes, and the limitations of current LLMs in capturing expert-level cognitive reasoning in chemistry.

\subsubsection{\textbf{The impact of model scaling.}} 

In our evaluation of two model sizes of LlaMA3, we found that LlaMA3-70B consistently outperforms LlaMA3-8B. The enhancement in performance on che-mical tasks with the increase in model parameters can be attributed to the augmented memory capacity and reasoning abilities of larger models. This allows for improved comprehension and detailed analysis of complex molecular structures and chemical phenomena, leading to notably superior performance in literature comprehension and scientific knowledge inference. 








\section{Discussion}

\subsection{Impact} 
ChemEval, introduced in this paper, aims to address the absence of benchmark in the domain of chemistry for LLMs by providing a comprehensive benchmark that encompasses a wide array of chemical tasks. Its strength lies in the inclusion of expert-reviewed data, which ensures a high level of authenticity and quality. The meticulous construction of ChemEval, supported by rigorous quality control measures and expert curation, positions it as a valuable platform for driving innovation and enhancement in chemical informatics and LLM evaluation. Its impact on the field is significant, offering a reliable and valid assessment of LLMs in the chemical domain. By providing a comparative analysis of model performance, ChemEval aids in the selection of suitable models for scientific research, thereby promoting the advancement of chemical science. Researchers can select large models based on the needs of actual scientific research, leveraging the models’ strengths to extract knowledge from scientific literature and experimental data, thereby promoting the advancement of scientific research.

\subsection{Limitations}
In our construction of chemical tasks, we observed several key findings: \textit{1). Prediction Challenges:} For tasks involving the prediction of spectral numerical features based on molecular structure descriptions, simulation of molecular dynamics behavior, and optimization of molecular 3D coordinates, all evaluated LLMs either produced refusals to respond or provided indiscriminate answers. \textit{2). Textual Description Limitation:} Despite their impressive capabilities in handling natural language tasks, the application of LLMs in the chemical domain is hindered by the current evaluation frameworks' reliance on textual descriptions. \textit{3). Integration Shortcomings:} This limitation is exacerbated by the lack of integration with professional molecular simulation tools, which are essential for accurate computational optimization and analysis. \textit{4). Domain-Specific Advantages:} General LLMs and chemistry-specific LLMs demonstrate distinct advantages across different tasks, underscoring the significance of domain-specific data and the challenge of catastrop-hic forgetting. \textit{5). Toxic Generation:} LLMs may generate content that is toxic, harmful or illegal, underscoring the necessity for stringent supervision of their generative processes.

\subsection{Future Work}
The future refinement of ChemEval, including the incorporation of multimodal tasks and advanced functionalities, will enhance its utility and applicability in the evolving landscape of AI and chemistry. We will invite experts to manually evaluate the results of the LLMs and compare them with the evaluation results of the paper. This will enhance the reliability of our evaluation system and make it more applicable to daily life and scientific research. In addition, research on agents has garnered significant attention recently\cite{huang2024understanding}, we aim to explore the integration of end-to-end agents to assist in chemical research endeavors in the future.








\section{Conclusion}

In this paper, we developed a comprehensive chemical evaluation system to assess the performance of popular LLMs across four levels of chemical tasks. The findings indicate that LLMs exhibit relatively poor performance on tasks requiring the understanding of molecular structures and scientific knowledge inference, whereas they perform better on tasks involving literature comprehension. This suggests both the potential for improvement and the need for further advancements in the application of LLMs to chemical tasks. Through this extensive evaluation, we demonstrate that there remains significant room for enhancement in the capabilities of LLMs across various chemical tasks. We hope our work will inspire future research to further explore and leverage the potential of LLMs in the field of chemistry. This has the potential to contribute to the transformation of scientific research paradigms and holds significant implications for the advancement of both the scientific community and artificial intelligence. Future work on ChemEval will integrate multimodal tasks and more sophisticated tasks and expert manual evaluations will be conducted to validate the result of ChemEval and other benchmarks to improve the evaluation system's dependability for practical and scientific applications.



\bibliographystyle{ACM-Reference-Format}  
\bibliography{ref}      




\appendix
\section{Supplementary Experimental Results}
\subsection{The experimental results for each level}
\label{Appendix A.1}
\subsubsection{The result of Advance Knowledge Answering}

Objective question answering encompasses the following tasks: \textit{Multiple Choice task (MCTask)}, \textit{Fill-in-the-Blank task (FBTask)}, and \textit{True/False Task (TFTask)}. For all the above tasks, only the answers need to be provided without any explanation, and the accuracy of the correct answers should be used as the model evaluation criterion. \textit{Subjective Questions include Short Answer Task (SATask)} and \textit{Calculation Task (CalcTask)}. The BLEU metric is employed to reflect the model's accuracy in generating relevant and precise molecular names for this task.

From the results, DeepSeek-V2 appears to be the best model for  Advance Knowledge Answering tasks. It demonstrates significantly superior performance in subjective tasks and achieves exceeding average performance in objective tasks. But other LLMs like GPT-4 and LLaMA3-70B demonstrate good performance in objective tasks and poor performance in subjective tasks. For \textit{True/Fal-se Task (TFTask)}, Claude-3.5 Sonnet achieves the best results. \textit{In Short Answer Task (SATask)} and \textit{Calculation Task (CalcTask)}, DeepS-eek-V2 consistently outperforms the others. We hypothesize that general LLMs have an advantage in this task because their pre-training encompasses extensive knowledge bases and they have acquired more complex reasoning and associative capabilities. Furthermore, advanced knowledge question answering does not require in-depth knowledge of molecular formulas and structures, allowing these models to perform better in this domain.
\begin{table*}[h!]
\Large
\caption{The result of Advance Knowledge Answering.}
\vspace{-0.2cm}
\resizebox{\textwidth}{!} {
\begin{tabular}{@{}cccccccccccccc@{}}
\toprule
\textbf{task} & \multicolumn{1}{l}{\textbf{Metric}} & \textbf{GPT-4} & \textbf{\begin{tabular}[c]{@{}c@{}}claude-3.5-\\ Sonnet\end{tabular}} & \textbf{\begin{tabular}[c]{@{}c@{}}ERNIE-4.0\end{tabular}} & \multicolumn{1}{l}{\textbf{Kimi}} & \textbf{\begin{tabular}[c]{@{}c@{}}LLaMA3- \\ 8B\end{tabular}} & \textbf{\begin{tabular}[c]{@{}c@{}}LLaMA3-\\  70B\end{tabular}} & \textbf{GLM-4} & \textbf{\begin{tabular}[c]{@{}c@{}}DeepSeek-\\ V2\end{tabular}} & \textbf{ChemDFM} & \textbf{\begin{tabular}[c]{@{}c@{}}LlaSMol-\\ Mistral\end{tabular}} & \textbf{\begin{tabular}[c]{@{}c@{}}ChemLLM-\\ 7B\end{tabular}} & \textbf{\begin{tabular}[c]{@{}c@{}}ChemSpark\end{tabular}} \\ \midrule
\multicolumn{14}{c}{\textit{\textbf{Objective Questions}}} \\ \midrule
MCTask & Acc & \textbf{70.00} & \textbf{70.00} & 65.00 & 60.00 & 45.00 & 65.00 & 55.00 & \textbf{70.00} & 50.00 & 15.00 & 0.00 & 60.00 \\
FBTask & Acc & 20.00 & 30.00 & \textbf{40.00} & 30.00 & 15.00 & 20.00 & 30.00 & 30.00 & 5.00 & 10.00 & 0.00 & 15.00 \\
TFTask & Acc & \textbf{90.00} & 80.00 & 85.00 & 65.00 & 80.00 & 90.00 & 70.00 & 80.00 & 70.00 & 40.00 & 10.00 & 85.00 \\ \midrule
\multicolumn{14}{c}{\textit{\textbf{Subjective Questions}}} \\ \midrule
SATask & BLEU & 10.53 & 13.71 & 14.81 & \multicolumn{1}{l}{9.41} & 5.12 & 6.01 & 12.01 & \textbf{45.15 } & 12.24 & 6.31 & 0.15 & 18.28 \\
CalcTask & Acc & 25.00 & \textbf{45.00} & 25.00 & 15.00 & 15.00 & 5.00 & 15.00 & \textbf{35.00} & 0.00 & 5.00 & 5.00 & 5.00 \\ \bottomrule
\end{tabular}
}
\label{tab:Answering}
\end{table*}

\begin{table*}[h!]
\Large
\caption{The result of Literature Understanding.}
\vspace{-0.2cm}
\resizebox{\textwidth}{!} {
\begin{tabular}{cccccccccccccc}
\toprule
\textbf{task} & \textbf{Metric} & \textbf{GPT-4} & \textbf{\begin{tabular}[c]{@{}c@{}}claude-3.5-\\ Sonnet\end{tabular}} & \textbf{ERNIE-4.0} & \textbf{Kimi} & \textbf{\begin{tabular}[c]{@{}c@{}}LLaMA3- \\ 8B\end{tabular}} & \textbf{\begin{tabular}[c]{@{}c@{}}LLaMA3-\\  70B\end{tabular}} & \textbf{GLM-4} & \textbf{\begin{tabular}[c]{@{}c@{}}DeepSeek-\\ V2\end{tabular}} & \textbf{ChemDFM} & \textbf{\begin{tabular}[c]{@{}c@{}}LlaSMol-\\ Mistral\end{tabular}} & \textbf{\begin{tabular}[c]{@{}c@{}}ChemLLM-\\ 7B\end{tabular}} & \textbf{\begin{tabular}[c]{@{}c@{}}ChemSpark\end{tabular}} \\ \hline
\multicolumn{14}{c}{\textit{\textbf{InfoE}}} \\ \hline
CNER & F1 & 63.76 & 76.73 & 75.43 & 61.80 & 76.09 & \textbf{77.56} & 51.59 & 66.19 & 58.99 & 19.44 & - & 75.08 \\
CERC & F1 & 24.97 & 31.02 & 24.92 & 21.43 & 22.58 & 24.28 & 23.33 & 22.40 & 12.26 & 3.40 & - & \textbf{40.18}  \\
SubE & Acc & 21.66 & 2.93 & 3.90 & 45.93 & 21.99 & 45.27 & \textbf{67.59} & 38.59 & 10.58 & 0 & 0 & 16.45 \\
AddE & F1 & 69.17 & \textbf{90.00} & 81.67 & 56.67 & 77.50 & 80.00 & 72.50 & 85.67 & 46.67 & 0 & - & 68.33 \\
SolvE & F1 & 85.00 & 85.00 & \textbf{90.00} & 85.00 & 89.00 & 89.00 & 74.00 & 89.00 & 82.50 & 0 & 15.00 & 85.43 \\
TempE & F1 & 10.00 & 70.00 & 60.00 & 15.00 & 65.00 & \textbf{80.00} & 5.00 & 70.00 & 20.00 & 0 & 15.00 & \textbf{80.00} \\
TimeE & F1 & 85.00 & \textbf{95.00} & 60.00 & 90.00 & 90.00 & 25.00 & \textbf{95.00} & 90.00 & 40.00 & 0 & 15.00 & 5.00 \\
ProdE & Acc & 25.69 & 4.39 & 5.78 & 65.05 & 43.98 & 62.50 & \textbf{76.39} & 36.80 & 35.41 & 0 & 0 & 32.41 \\
CharME & F1 & 54.20 & \textbf{74.36} & \textbf{69.13} & 30.70 & 43.40 & 59.17 & 35.00 & 65.40 & 26.20 & 0 & - & 15.00 \\
CatTE & F1 & 90.00 & \textbf{100.00} & 80.00 & 90.00 & 85.00 & 95.00 & 90.00 & \textbf{95.00} & 55.00 & 0 & 60.00 & 25.00 \\
YieldE & F1 & \textbf{45.00} & 25.00 & 35.00 & \textbf{45.00} & 20.00 & 25.00 & 35.00 & 35.00 & 20.00 & 0 & 20.00 & 20.00 \\ \hline
\multicolumn{14}{c}{\textit{\textbf{InducGen}}} \\ \hline
AbsGen & BLEU-2 & 64.17 & 59.58 & - & 67.18 & \textbf{67.31} & 60.66 & 61.23 & 65.22 & 53.97 & 0.31 & 0 & 3.5 \\
OLGen & BLEU-2 & 2.95 & 6.41 & - & \textbf{60.10} & 0 & 1.77 & 10.62 & 0.37 & 22.43 & 0.78 & 0 & 3.11 \\
TopC & Acc & 55.00 & 60.00 & 50.00 & 55.00 & 25.00 & \textbf{70.00} & 50.00 & 55.00 & 40.00 & 25.00 & 35.00 & 20.00 \\
ReactTR & F1 & 20.00 & \textbf{56.76} & 30.00 & 20.00 & 15.00 & 45.00 & 25.00 & 20.00 & 15.00 & 0 & 25.00 & 25.00 \\ \bottomrule
\end{tabular}
}
\vspace{-0.2cm}
\label{tab:Literature}
\end{table*}

\subsubsection{The result of Literature Understanding}

Information Extraction is subdivided into the following areas: \textit{Chemical Named Entity Recognition (CNER)}, \textit{Chemical Entity Relationship Classification (CERC)}, \textit{Synthetic Reaction Substrate Extraction (SubE)}, \textit{Synthetic Reaction Additive Extraction (AddE)}, \textit{Synthetic Reaction Solvent Extraction (SolvE)}, \textit{Reaction Temperature Extraction (TempE)}, \textit{Reaction Time Extraction (TimeE)}, \textit{Reaction Product Extraction (ProdE)}, \textit{Characterization Method Extraction (CharME)}, \textit{Catalysis Type Extraction (CatTE)}, and \textit{Yield Extraction (YieldE)}. All the above sub-tasks use accuracy as the evaluation metric.  Inductive Generation incorporates \textit{Chemical Paper Abstract Generation (AbsGen)}, \textit{Research Outline Generation (OLGen)}, \textit{Chemical Literature Topic Classification (TopC)}, and \textit{Reaction Type Recognition and Induction (ReactTR)}. AbsGen and OLGen two sub-tasks use BLEU-2 as the evaluation metric. TopC and ReactTR two sub-tasks use accuracy as the evaluation metric.

Analyzing the results of Chemical Literature Comprehension, LlaMA3 70B performs the best, taking into account accuracy, F1 and BLEU score. LLaMA3 70B achieves. LLaMA3 70B realize the best outcomes in five tasks, including \textit{Chemical Named Entity Recognition (CNER)}, \textit{Reaction Temperature Extraction (TempE)}, \textit{Catalysis Type Extraction (CatTE)}, \textit{Chemical Literature Topic Classification (TopC)} and \textit{Reaction Type Recognition and Induction (ReactTR)}. In \textit{Chemical Paper Abstract Generation (AbsGen)} and \textit{Research Outline Generation (OLGen)}, especially in OLGen, Kimi achieves a 60.1$\%$ result, showing remarkable performance advantages over other models. General-purpose large models also achieved better results in this task. This can be attributed to the fact that document understanding is a critical evaluation criterion for these models. During pre-training, general-purpose models are specifically trained to enhance their document comprehension capabilities. Due to their large parameter size, they can better understand and generate natural language text. Consequently, in the task of chemical literature comprehension, which requires advanced language understanding and generation, these models exhibit superior performance. In addition, the extremely poor performance of LlaSMol and ChemLLM highlights the challenge in the process of instruction fine-tuning.

\begin{table*}[h!]
\Large
\caption{The result of Molecular Understanding}
\vspace{-0.2cm}
\resizebox{\textwidth}{!} {
\begin{tabular}{cccccccccccccc}
\toprule
\textbf{task} & \textbf{Metric} & \textbf{GPT-4} & \textbf{\begin{tabular}[c]{@{}c@{}}claude-3.5-\\ Sonnet\end{tabular}} & \textbf{ERNIE-4.0} & \textbf{Kimi} & \textbf{\begin{tabular}[c]{@{}c@{}}LLaMA3- \\ 8B\end{tabular}} & \textbf{\begin{tabular}[c]{@{}c@{}}LLaMA3-\\  70B\end{tabular}} & \textbf{GLM-4} & \textbf{\begin{tabular}[c]{@{}c@{}}DeepSeek-\\ V2\end{tabular}} & \textbf{ChemDFM} & \textbf{\begin{tabular}[c]{@{}c@{}}LlaSMol-\\ Mistral\end{tabular}} & \textbf{\begin{tabular}[c]{@{}c@{}}ChemLLM-\\ 7B\end{tabular}} & \textbf{\begin{tabular}[c]{@{}c@{}}ChemSpark\end{tabular}} \\ \hline
\multicolumn{14}{c}{\textit{\textbf{MNGen}}} \\ \hline
MolNG & BLEU & 40.21 & 40.8 & 44.99 & 8.86 & 13.85 & 43.11 & 23.89 & 24.43 & 75.91 & 75.65 & 0.94 & \textbf{80.00} \\ \hline
\multicolumn{14}{c}{\textit{\textbf{MNTrans}}} \\ \hline
IUPAC2MF & Exact Match & 5.00 & 20.00 & 30.00 & 5.00 & 0.00 & 15.00 & 10.00 & 0 & 25.00 & 0.00 & 0 & \textbf{70.00} \\
SMILES2MF & Exact Match & 0 & 5.00 & 5.00 & 5.00 & 0 & 0 & 0 & 0 & 45.00 & 0 & 0 & \textbf{75.00} \\
IUPAC2SMILES & Exact Match & 0 & 25.00 & 15.00 & 0 & 0 & 0 & 0 & 0 & 10.00 & 65.00 & 0 & \textbf{80.00} \\
SMILES2IUPAC & Exact Match & 0 & 0 & 0 & 0 & 0 & 0 & 0 & 0 & 0 & 5.00 & 0 & \textbf{45.00} \\
S2S & Exact Match & 0 & 0 & 5.00 & 0 & 0 & 0 & 5.00 & 0 & 0 & 0 & 0 & \textbf{85.00} \\ \hline
\multicolumn{14}{c}{\textit{\textbf{MPP}}} \\ \hline
MolPC & {\color[HTML]{545454} Accuracy} & 56.50 & 53.60 & 7.65 & 55.05 & 51.60 & 44.75 & 48.10 & 55.55 & 60.95 & 39.00 & 26.00 & \textbf{76.75} \\
MolPR & {\color[HTML]{545454} Rank} & 6.00 & \textbf{2.86} & 5.43 & 8.57 & 6.14 & 4.29 & 6.86 & 7.71 & 8.57 & 8.14 & 9.71 & 4.00
 \\ \hline
\multicolumn{14}{c}{\textit{\textbf{MolDesc}}} \\ \hline
Mol2PC & BLEU-2 & 13.87 & 19.74 & 28.98 & 47.28 & 35.54 & 39.64 & 41.97 & 40.86 & 38.86 & \textbf{52.78} & 0 & 50.44 \\ \bottomrule
\end{tabular}
}
\label{tab:Molecular}
\end{table*}

\begin{table*}[h!]
\Large
\caption{The result of Scientific Knowledge Deduction. }
\vspace{-0.2cm}
\resizebox{\textwidth}{!} {
\begin{tabular}{cccccccccccccc}
\toprule
\textbf{task} & \textbf{Metric} & \textbf{GPT-4} & \textbf{\begin{tabular}[c]{@{}c@{}}claude-3.5-\\ Sonnet\end{tabular}} & \textbf{ERNIE-4.0} & \textbf{Kimi} & \textbf{\begin{tabular}[c]{@{}c@{}}LLaMA3- \\ 8B\end{tabular}} & \textbf{\begin{tabular}[c]{@{}c@{}}LLaMA3-\\  70B\end{tabular}} & \textbf{GLM-4} & \textbf{\begin{tabular}[c]{@{}c@{}}DeepSeek-\\ V2\end{tabular}} & \textbf{ChemDFM} & \textbf{\begin{tabular}[c]{@{}c@{}}LlaSMol-\\ Mistral\end{tabular}} & \textbf{\begin{tabular}[c]{@{}c@{}}ChemLLM-\\ 7B\end{tabular}} & \textbf{\begin{tabular}[c]{@{}c@{}}ChemSpark\end{tabular}} \\ \hline
\multicolumn{14}{c}{\textit{\textbf{ReSyn}}} \\ \hline
SubRec & Acc & 0 & 7.77 & 1.90 & 4.44 & 0 & 0 & 0 & 0 & 8.22 & 0 & - & \textbf{20.00} \\
PathRec & Acc & 22.50 & \textbf{32.5} & 17.50 & 17.50 & 5.00 & 22.50 & 12.50 & 25.00 & 15.00 & 2.50 & 0 & 12.50 \\
SynDE & RMSE(Valid Num)  & - (0)& - (0)& - (0)& - (0)& - (0)& 4.35(20) & - (0) & - (0) & - (0) & 2.15(10) & 45.81(13) & \textbf{2.04(19)} \\ \hline
\multicolumn{14}{c}{\textit{\textbf{RCRec}}} \\ \hline
LRec & Acc & 18.18 & 5.00 & 23.87 & 18.18 & 6.81 & 4.55 & 16.70 & 17.27 & 20.13 & 0 & 15.00 & \textbf{45.00} \\
RRec & Acc & 34.17 & 17.85 & 34.67 & 34.17 & 12.06 & 27.51 & 28.33 & 25.33 & 19.00 & 0 & - & \textbf{51.76} \\
SolvRec & Acc & \textbf{50.00} & 15.00 & \textbf{50.00} & 42.50 & 25.42 & 24.58 & 48.75 & \textbf{50.00} & 5.00 & 0 & 5.00 & 40.00 \\
CatRec & Acc & 0 & 0 & 0 & 0 & 0 & 0 & 0 & 0 & 0 & 0 & - & 0 \\
TempRec & RMSE(Valid Num) & 171.93(20) & \textbf{16.73(20)} & 28.83(20) & 25.79(20) & 27.75(20) & 33.03(20) & 35.24(20) & 234.47(18) & 71.48(20) & 76.14( 20) & 73.58( 16 ) & \textbf{16.73 (20) } \\
TimeRec & RMSE(Valid Num) & 10.67(20) & 23.43(20) & 11.31(18) & 23.56(15) & 16.17(20) & 13.62(20) & 16.93(20) & 36.30(17) & 17.81(20) & 9.41(17) & 98.44( 19) & \textbf{9.09(20)} \\ \hline
\multicolumn{14}{c}{\textit{\textbf{ROP}}} \\ \hline
PPre & Acc & 0 & 0 & 0 & 0 & 3.33 & 0 & 0 & 0 & 3.33 & 35.00 & 0 & \textbf{60.00} \\
YPred & Acc & 70.00 & 20.00 & 35.00 & \textbf{85.00} & 35.00 & 45.00 & 20.00 & 60.00 & 20.00 & 15.00 & 45.00 & 65.00 \\
RatePred & Overlap & 14.54 & \textbf{23.50} & 11.25 & 17.06 & 8.42 & 5.97 & 6.70 & 10.77 & 4.71 & 0 & 5.44 & 4.21 \\ \hline
\multicolumn{14}{c}{\textit{\textbf{RMA}}} \\ \hline
IMDer & Acc & 25.00 & 35.00 & 30.00 & 30.00 & 25.00 & 25.00 & \textbf{40.00} & 35.00 & 10.00 & 20.00 & 10.00 & 5.00 \\ \bottomrule
\end{tabular}
}
\vspace{-0.2cm}
\label{tab:Knowledge}
\end{table*}

\subsubsection{The result of Molecular Understanding}

Molecular name generation only contains one subtask, Molecular Name Generation from \textit{Text Description (MolNG)}. The BLEU metric is used to measure the accuracy of the generated molecular names. Molecular Name Translation includes \textit{IUPAC to Molecular Formula (IUPAC2MF)}, \textit{SMI-LES to Molecular Formula (SMILES2MF)}, \textit{IUPAC to SMILES (IUPAC2-SMILES)}, \textit{SMILES to IUPAC (SMILES2IUPAC)}, and \textit{SMILES to SELFIES and SELFIES to SMILES Translation (S2S)}. All five subtasks are evaluated using an Exact Match criterion. Molecular Property Prediction is encapsulated in two core objectives: \textit{Molecule Property Classification (MolPC)}, which predicts categories of properties such as ClinTox, HIV inhibition, BBBP, SIDER effects, and polarity; and \textit{Molecule Property Regression (MolPR)}, focusing on estimating numerical values related to HOMO, LUMO, ESOL, Lipophilicity, polarity, melting point, and boiling point. Assessment of classification tasks is grounded in accuracy metrics. The effectiveness of regression tasks is evaluated using the RMSE metric and is ranked accordingly. Molecular Description encompasses only one subtask, Physicochemical Property Prediction from \textit{Molecular Structure (Mo-l2PC)}. Performance in Mol2PC is assessed using the BLEU-2 score.

In the results for the molecular understanding task, ChemSpark achieved the best results in the \textit{Molecular Name Generation (MNGen)} and \textit{Molecular Name Translation (MNTrans)} tasks. Regarding the performance of models in molecular name translation, almost all general models and most chemistry-specific models still failed to complete this task. Although these models can generate names in a standard format, they fail to provide completely correct answers. This indicates that while the models understand the text format, they do not correctly understand the molecules. However, ChemSpark demonstrated a significant performance advantage compared to other models, likely due to its exposure to a broader range of chemical problem types during the fine-tuning phase. For the \textit{Molecular Property Prediction (MPP)} and \textit{Molecular Description (MolDesc)} tasks, chemistry-specific large models exhibited a greater advantage due to the deeper nature of these tasks, which involve complex associations between molecular structure and chemical properties. In contrast, general models performed poorly due to a lack of domain-specific knowledge.

\subsubsection{The result of Scientific Knowledge Deduction}
Retrosynthetic Analysis comprises \textit{Substrate Recommendation (SubRec)}, \textit{Synthetic Pathway Recommendation (PathRec)} and \textit{Synthetic Difficulty Evaluation (SynDE)}. SubRec is measured by F1 metric in predicting the correct substrates.  The effectiveness of PathRec is evaluated by the accuracy of the recommended pathways. SynDE is gauged using the RMSE as the evaluation metric. \textit{Reaction condition recommendation integrates Ligand Recommendation (LRec)}, \textit{Reagent Recommendation (RRec)}, \textit{Solvent Recommendation (SolvRec)}, \textit{Catalyst Recommendation (CatRec)}, \textit{Reaction Temperature Recommendation (TempRec)}, and \textit{Reaction Time Recommendation (TimeRec)}, each targeting a specific component of the reaction condition optimization. LRec, RRec, SolvRec and CatRec use F1 metric as the evaluation metric, while TempRec and TimeRec use RMSE as the evaluation metric. Reaction outcome prediction encompasses \textit{Reaction Product Prediction (PPred)}, \textit{Product Yield Prediction (YPred)}, and \textit{Reaction Rate Prediction (RatePred)}. PPred utilizes F1 score, YPred employs Accuracy, and RatePred involves activation energy characterization using overlap as evaluation metrics. Reaction mechanism analysis encompasses only one subtask: \textit{Intermediate Derivation(IMDer)}. This task uses Accuracy as the evaluation metric.

For scientific knowledge deduction tasks, ChemSpark consistently outperformed other models in substrate recommendation, synthesis difficulty estimation, and various reaction condition recommendations. This superior performance is attributed to its comprehensive fine-tuning on chemical-specific data, enabling ChemS-park to better understand complex chemical relationships and patterns. In contrast, general-purpose models like Claude-3.5 Sonnet and GPT-4 excelled in tasks such as pathway recommendation and intermediate detection, owing to their extensive knowledge bases and strong reasoning capabilities. Additionally, we observed that, except for ChemSpark, the prediction accuracy for tasks like \textit{substrate recommendation (SubRec)} and \textit{ligand recommendation (LRec)} was nearly zero for other models. A thorough examination of the datasets used for fine-tuning these large chemical models revealed that, except for ChemSpark, they did not include similar tasks, resulting in a lack of relevant knowledge within the models. For \textit{catalyst recommendation (CatRec)}, no model achieved any accuracy, indicating a general difficulty in recommending catalysts due to the complexity and specificity of catalytic processes. Despite the large parameter sizes of these general models, their lack of specialized training for tasks requiring deep chemical knowledge led to poor performance in highly specialized tasks. This underscores the value of aligning model training with specific task requirements to achieve optimal performance.

\begin{table*}[h!]
\Large
\caption{Comparative Analysis of 3-Shot Performance Among Mainstream LLMs Across Four Levels of Chemical Evaluation}
\vspace{-0.2cm}
\resizebox{\textwidth}{!} {
\begin{tabular}{ccccccccccccc}
\hline
Metric & \textbf{GPT-4} & \textbf{\begin{tabular}[c]{@{}c@{}}claude-3.5-\\ Sonnet\end{tabular}} & \textbf{ERNIE-4.0} & \textbf{Kimi} & \textbf{\begin{tabular}[c]{@{}c@{}}LLaMA3- \\ 8B\end{tabular}} & \textbf{\begin{tabular}[c]{@{}c@{}}LLaMA3- \\  70B\end{tabular}} & \textbf{GLM-4} & \textbf{\begin{tabular}[c]{@{}c@{}}DeepSeek-\\ V2\end{tabular}} & \textbf{ChemDFM} & \textbf{\begin{tabular}[c]{@{}c@{}}LlaSMol-\\ Mistral\end{tabular}} & \textbf{\begin{tabular}[c]{@{}c@{}}ChemLLM-\\ 7B\end{tabular}} & \textbf{\begin{tabular}[c]{@{}c@{}}ChemSpark\end{tabular}} \\ \hline
\multicolumn{13}{c}{\textit{\textbf{Advanced Knowledge Question Answering}}} \\ \hline
Accuracy & 50 & 53.75 & \textbf{58.75} & 42.50 & 35.00 & 50.00 & 42.50 & 50.00 & 32.50 & 8.75 & 6.25 & 40.00 \\
BLEU-2 & 15.65 & 14.19 & 20.9 & 18.07 & 8.18 & 10.19 & 17.28 & 36.55 & 11.41 & 0.08 & 0.15 & \textbf{26.82} \\ \hline
\multicolumn{13}{c}{\textit{\textbf{Literature Understanding}}} \\ \hline
F1 & 67.14 & 73.45 & 70.80 & 61.31 & 56.83 & \textbf{71.39} & 63.85 & 70.33 & 38.47 & 2.06 & - & 58.91 \\
Accuracy & \textbf{66.71} & 20.88 & 24.49 & 55.73 & 46.19 & 43.39 & 63.04 & 49.69 & 22.95 & 0 & 15.00 & 29.96 \\
BLEU-2 & - & - & - & - & - & - & - & - & - & - & - & - \\ \hline
\multicolumn{13}{c}{\textit{\textbf{Molecular Understanding}}} \\ \hline
BLEU & 61.31 & 62.30 & 50.39 & 18.46 & 12.65 & 61.85 & 18.68 & 12.07 & 55.23 & 0 & 0 & \textbf{72.84} \\
Exact Match & 3.00 & 8.00 & 8.00 & 0 & 0 & 0 & 0 & 2.00 & 7.00 & 0 & 0 & \textbf{60.00} \\
Accuracy & 58.3 & 63.75 & 67.05 & 62.20 & 56.15 & 54.25 & 51.70 & 57.50 & 54.05 & 38.00 & 18.00 & \textbf{75.30} \\
Rank & 4.71 & \textbf{2.00} & 4.86 & 7.43 & 5.86 & 4.14 & 6.57 & 8.00 & 9.57 & 9.86 & 10.00 & 4.71 \\
BLEU-2 & 26.08 & 43.16 & 39.06 & 53.62 & 44.47 & 52.62 & 41.85 & 50.12 & 24.59 & 0.91 & 0 & \textbf{54.17} \\ \hline
\multicolumn{13}{c}{\textbf{Scientific Knowledge Deduction}} \\ \hline
F1 & 17.48 & 12.22 & 31.43 & 18.17 & 28.25 & \textbf{38.89} & 15.59 & 23.74 & 14.55 & 0 & - & 32.97 \\
Accuracy & 42.5 & 30.00 & 40.00 & \textbf{48.33} & 35.00 & 38.33 & 20.83 & 46.67 & 29.17 & 0 & 1.67 & 35.00 \\
RMSE(Valid Num) & 8.32(60)& 9.49(60) & 14.26(45) & 19.35(49) & 7.55(60) & 10.48(60) & 12.95(60) & 7.91(59) & 83.73(51) & 27.79(56) & 29.94(52) & 7.93(54) \\
Overlap & 15.89 & 19.30 & 22.49 & 18.85 & 18.02 & \textbf{19.35} & 15.59 & 17.64 & 15.77 & 9.05 & 11.32 & 12.42 \\ \bottomrule
\end{tabular}
}
\vspace{-0.2cm}
\label{tab:3shot}
\end{table*}

\subsection{3-shot performance}
\label{Appendix A.2}
Under the few-shot setting, GPT-4 and ERNIE-4.0 exhibit significant performance enhancements in objective question answering, while other models remain largely unchanged. In the domain of chemical literature comprehension and objective question answering, many models also achieve improved results, indicating enhanc-ed contextual learning capabilities. However, the few-shot setting adversely affects the performance of ChemDFM, LlaSMol, and Che-mLLM, suggesting that specialized model fine-tuning sacrifices in-context learning abilities. Regarding molecular understanding and scientific knowledge deduction, the improvements across all models are not particularly significant, and many metrics even show a decline.

\section{Prompt Examples}
\label{appendixb}

The following section presents a set of {\benchmark} prompts designed to evaluate a chemical language model. These prompts are categorized under four primary indicators: Advanced Knowledge Questions, Literature Comprehension, Molecular Understanding, and Scientific Reasoning. Each primary indicator is further divided into secondary and tertiary indicators. The examples provided here are zero-shot examples, where the model is required to generate responses without any prior examples or training on similar tasks.

Pay attention that some of the tasks below are in Chinese: all sub-tasks of Advanced Knowledge Question Answering, Spectral Feature Prediction from Molecular Structure, Molecular 3D Coordinate Optimization and Reaction Rate Prediction. Other tasks are all in English. To facilitate understanding for non-Chinese speaking audiences, we provide the English translations of these prompts, which are marked with \textbf{(Translation)}. Note that the actual input used was in Chinese, and the following English text is provided for reference only. 
\subsection{Advanced Knowledge Questions}

\subsubsection{Objective Question-Answering}

\paragraph{Multiple Choice task(MCTask):} In this assessment method, participants are tasked with selecting the correct answer from a provided list of options. Commonly used to gauge knowledge retention and comprehension, the MCT requires respondents to indicate their choice using the designated letter format (e.g., A, B, C, D) without additional commentary. The focus is on accuracy within the strict confines of the given format.

\begin{tcolorbox}[title=Prompt Example]
    \textbf{Query(Translation):}
    Given a chemistry problem and multiple choices, please select the correct answer. Your answer should be one of `A', `B', `C', `D', etc. You only need to output the correct answer's letter without any additional explanation. The problem and options are as follows:\\
    Which of the following methods can be used to prepare a cis-alkene from an alkyne?\\
    A. Sodium borohydride reduction B. Hydroboration-reduction C. Lithium aluminum hydride reduction D. Alkali metal and liquid ammonia reduction \\
    You must output your judgment and strictly follow the output format, \{''answer'': ''your answer''\}. I don't need any explanations, you only need to output your judgment in the specified format.

    \vspace{3mm}
    
    \textbf{Answer:} ''B'',
\end{tcolorbox}

\paragraph{Fill-in-the-Blank task(FBTask):} This task assesses the LLMs' recall of specific chemical terms or concepts by requiring participants to complete statements or sentences with the appropriate term or phrase. Indicated by underscores (`\_\_\_'), these blanks must be filled in correctly according to the specified guidelines.

\begin{tcolorbox}[title=Prompt Example]

    \textbf{Query(Translation):}
    Given a chemistry fill-in-the-blank question, the `\_\_\_' underline area is where you need to fill in the answer. You only need to output the correct answer for each `\_\_\_' underline area. The requirements are as follows:\\
    1) Do not repeat the content of the question;\\
    2) Do not output any analysis process;\\
    3) If there are multiple underline areas, please use the following format:\\
    1, 2, ...\\
    The question is as follows:\\
    The selection of a distillation flask is based on the criterion that the liquid volume should occupy \_\_\_\_\_ of the flask volume. When the boiling point of the distillate is below $80^{\circ}\mathrm{C}$, use \_\_\_\_\_ heating. When the boiling point is between $80-200^{\circ}\mathrm{C}$, use \_\_\_\_\_ heating. Do not use \_\_\_\_\_ for direct heating.\\
    You must output your judgment and strictly follow the output format, \{''answer'': ''your answer''\}. I don't need any explanations, you only need to output your judgment in the specified format.

    \vspace{3mm}
    
    \textbf{Answer(Translation):} ''1. 1/3 - 2/3; 2. water bath; 3. oil bath; 4. electric heating mantle''
\end{tcolorbox}

\paragraph{True/False Task(TFTask):} In this task, the LLMs' ability to assess the accuracy of statements is tested through binary judgments. Participants evaluate given statements, responding with either "True" if the statement is correct, or "False" if it is incorrect.
\begin{tcolorbox}[title=Prompt Example]

    \textbf{Query(Translation):}
    Given a chemistry true or false question, if you think the statement is correct, output "true"; otherwise, output "false." You only need to output "true" or "false" without any additional explanation.\\
    The question is as follows:\\
    All trans isomers of alkenes belong to the E-type.\\
    You must output your judgment and strictly follow the output format, \{''answer'': ''your answer''\}. I don't need any explanations, you only need to output your judgment in the specified format.
                    
    \vspace{3mm}
    
    \textbf{Answer(Translation):} ''False''
\end{tcolorbox}

\subsubsection{Subjective Question-Answering}

\paragraph{Short Answer Task(SATask):} In this task, participants demonstra-te their grasp of concepts or their skill in summarizing information succinctly. Respondents are instructed to answer chemistry-related questions in a brief, structured manner, adhering to a sequential format (e.g., 1. Answer to question 1, 2. Answer to question 2), ensuring clarity and conciseness in their responses.

\begin{tcolorbox}[title=Prompt Example]

    \textbf{Query(Translation):}
    Given a chemistry short answer question, please answer the questions in order. The format should be 1. answer to question 1, 2. answer to question 2, and so on.\\
    The question is as follows:\\
    What should be noted during the fractional distillation operation?\\
    You must output your judgment and strictly follow the output format, \{''answer'': ''your answer''\}. I don't need any explanations, you only need to output your judgment in the specified format.

    \vspace{3mm}
    
    \textbf{Answer(Translation):}
    ''(1) When assembling the apparatus, the fractionating column should be as vertical as possible to ensure that the condensate from above and the rising vapor from below can fully exchange heat and mass, thus improving separation efficiency. (2) Select a suitable heating bath based on the boiling point range of the fractionated liquid, and avoid direct flame heating on an asbestos net. Heat the bath gently so that the bath temperature rises slowly and uniformly. (3) When the liquid starts to boil and vapor enters the fractionating column, adjust the bath temperature to ensure that the vapor rises slowly and uniformly along the column walls. If the room temperature is low or the liquid has a high boiling point, wrap the fractionating column with asbestos rope or glass wool to reduce heat loss. (4) When the vapor reaches the top of the column and liquid begins to distill, carefully adjust the bath temperature to control the distillation rate to one drop every 2-3 seconds. If the distillation rate is too fast, product purity decreases; if too slow, the vapor will rise intermittently, causing fluctuations in distillation temperature. (5) Collect fractions according to experimental requirements and weigh each fraction at the end of the experiment.''
\end{tcolorbox}

\paragraph{Calculation Task(CalcTask):} This task evaluates LLMs' quantitative skills and their ability to apply chemical principles through solving problems that require mathematical operations. Participan-ts must submit their calculated solutions in the prescribed format.

\begin{tcolorbox}[title=Prompt Example]

    \textbf{Query(Translation):}
    Given a chemistry calculation question, please answer the questions in order. The format should be 1. answer to question 1, 2. answer to question 2, and so on.\\
    The question is as follows:\\
    At a certain temperature, the rate constant for the reaction \(2 \mathrm{NO}(\mathrm{g}) + \mathrm{O}_{2}(\mathrm{g}) \rightarrow 2 \mathrm{NO}_{2}(\mathrm{g})\) is \(k = 8.8 \times 10^{-2} \, \mathrm{dm}^{6} \cdot \mathrm{mol}^{-2} \cdot \mathrm{s}^{-1}\). It is known that this reaction is first order with respect to \(\mathrm{O}_{2}\). What is the reaction rate when the concentrations of both reactants are \(0.050 \, \mathrm{mol} \cdot \mathrm{dm}^{-3}\)?\\
    You must output your judgment and strictly follow the output format, \{''answer'': ''your answer''\}. I don't need any explanations, you only need to output your judgment in the specified format.

    \vspace{3mm}
    
    \textbf{Answer(Translation):}
''From the unit of the rate constant \(k\) (\(\mathrm{dm}^{6} \cdot \mathrm{mol}^{-2} \cdot \mathrm{s}^{-1}\)), it can be seen that the reaction 
\(2 \mathrm{NO}(\mathrm{g}) + \mathrm{O}_{2}(\mathrm{g}) \rightarrow 2 \mathrm{NO}_{2}(\mathrm{g})\) is a third-order reaction.
Since the reaction is first order with respect to \(\mathrm{O}_{2}\), it must be second order with respect to \(\mathrm{NO}\). The rate equation is therefore 
\(v = k c(\mathrm{O}_{2})[c(\mathrm{NO})]^{2}\).
Substituting the given concentrations and the rate constant \(k\) into the rate equation, we get
\(v = 8.8 \times 10^{-2} \, \mathrm{dm}^{6} \cdot \mathrm{mol}^{-2} \cdot \mathrm{s}^{-1} \times 0.05 \, \mathrm{mol} \cdot \mathrm{dm}^{-3} \times \left(0.05 \, \mathrm{mol} \cdot \mathrm{dm}^{-3}\right)^{2} = 1.1 \times 10^{-5} \, \mathrm{mol} \cdot \mathrm{dm}^{-3} \cdot \mathrm{s}^{-1}\).''
\end{tcolorbox}

\subsection{Chemical Literature Comprehension}

\subsubsection{Information Extraction}
\paragraph{Chemical Named Entity Recognition(CNER):} This task involves identifying and classifying chemical entities within text, such as compounds, elements, or molecular structures. 

\begin{tcolorbox}[title=Prompt Example]

    \textbf{Query:} You are an expert in the field of chemistry. Find and list all the instances of the chemical entities in the following content.\\
    The text of the material is shown below. Absorbance decreased with the increase of methanol concentration in sample solution from 2 \% to 40 \% , while the standard curve became most linear when the sample solution contained 10 \% methanol .\\
    You need to output all the chemical entities mentioned in the material text (separated by '', ''), and follow the following output format strictly: \\
    \{''answer'': '' Chemical entity 1, Chemical entity 2, Etc''\}

    \vspace{3mm}
    
    \textbf{Answer:} ''methanol, methanol''
\end{tcolorbox}

\paragraph{Chemical Entity Relationship Classification(CERC):} In this task, the model is required to classify the relationships between identified chemical entities, focusing specifically on extracting chemical-disease associations from the provided texts. The goal is to accurately identify and format pairs of chemical and disease entities that exhibit relationships such as 'is a precursor of' or 'reacts with'.

\begin{tcolorbox}[title=Prompt Example]

    \textbf{Query:} You are an expert in the field of chemistry. You are provided with a set of clinical trial summaries. Extract the chemical-disease relations from the summaries and present your findings in the format of (Subject, Object).\\
    Eating disorders and the associated behavioural problems and drug abuse are uncommon in pregnancy. When they do occur they are often unrecognized because of denial but when significant may pose a risk to both the mother and her fetus. This case illustrates a number of problems that may be encountered in women with eating disorders in pregnancy, including prolonged and recurrent metabolic disturbances and diuretic abuse. In particular, it illustrates the derangements of thyroid function seen in pregnant women with eating disorders and reminds us that when a cause for thyrotoxicosis remains obscure, thyroxine abuse should be considered and explored.\\
    You need to output all chemical and disease entity pairs (separated by '', '') mentioned in the input text, following the following output format to the letter:\\
    \{''answer'': '' (Chemical 1, disease 1), (chemical 2, disease 2), Etc''\}

    \vspace{3mm}
    
    \textbf{Answer:} ''(thyroxine, thyrotoxicosis)''
\end{tcolorbox}

\paragraph{Synthetic Reaction Substrate Extraction(SubE):} This task entails the model performing named entity recognition to identify and label substrates—crucial starting materials—that contribute heavy atoms to the product in chemical synthesis reactions, as described in the literature.

\begin{tcolorbox}[title=Prompt Example]

    \textbf{Query:} You are an expert in the field of chemistry and will be required to perform named entity recognition (ner) task. The following paragraph is a string list of a sentence of a paper from which you need to combine the information in the sentence to extract the important entities and give them entity labels. For each element of the sentence list, an entity label is required. \\
    The permitted identification labels are B-type, I-type, O, where type contains Reactants (Chemical substances that contribute heavy atoms to the substance) product), Catalyst\_Reagents (Chemical substances that participate in the reaction but do not contribute heavy atoms (e.g., acid, base,  metal complexes)), Workup\_reagents (Chemical substances that are used after the reactions to terminate the reactions or obtain the products (e.g., quenching reagents, extraction solvent,  neutralizing acids/bases)), Solvent (Chemical substances that are used to dissolve/mix other chemicals, typically quantified by volume and used in superstoichiometric amounts (e.g., water, toluene,  THF)), Temperature (Temperature at which the reaction occurs), Time (Duration of the reaction performed), Reaction (Descriptions about the type of chemical reaction) and Yield (Yield of the product). \\
    Note: B means the beginning of the entity, I means inside the entity, and O means not belonging to any entity. All possible identifiers include: O,B-Reactants, I-Reactants, B-Catalyst\_Reagents, I-Catalyst\_Reagents, B-Workup\_reagents, I-Workup\_reagents, B-Reaction,  I-Reaction, B-Solvent, I-Solvent, B-Yield, I-Yield, B-Temperature, I-Temperature, B-Time, I-Time. \\
    You need to determine the type of each element in the sentence list according to the entity type above and fill the corresponding entity label in the output list, even if the element in the list is not a word, you need to determine and output. You must output your prediction, i.e. a list, and follow the output format exactly as follows:\{''answer'': ''a list ''\}. Next is the sentence list: \\ 
    \text{[}`Alkylation', `with', `ethyl', `3,3-diamino-2-propenoatelg', `(', `4', `)', `(', `Scheme', `IA', `)', `yields', `1', `in', `58', `\%', `yield', `from', `2.'] Please answer in the following format strictly: [a, b, c... ](where a, b, c represent possible labels); Do not output any explanatory text, only output the list. please ensure the output list length matches the input list length.

    \vspace{3mm}
    
    \textbf{Answer:} ''[`B-Reaction', `O', `B-Reactants', `I-Reactants', `O', `B-Reactants', `O', `O', `O', `O', `O', `O', `B-Prod', `O', `O', `O', `O', `O', `O']''
\end{tcolorbox}

\paragraph{Synthetic Reaction Additive Extraction(AddE):} This task centers on recognizing additives in chemical synthesis reactions from textual data. Models are required to accurately identify and format information about these substances, which play a role in influencing the reaction without being fully consumed.

\begin{tcolorbox}[title=Prompt Example]

    \textbf{Query:} You are an expert in the field of chemistry and here is a task to extract information from the chemistry literature that requires you to extract the Addictives (Op) involved from the input.\\
    You must output your prediction, i.e. an effective chemical agent, and follow the output format exactly as follows:\\
    \{''answer'': ''an effective chemical agent ''\}. I don't need any explanation, you just need to output your judgment in format.\\
    The contents of input are as follows:\\
    ''input'':''Reaction conditions: 1 (1.5 equiv), 2a (0.25 mmol), 3a (1.0 mL), Bu4NBF4 (0.1 mmol), MeCN (9.0 mL), carbon rod anode, nickel plate cathode, undivided cell, constant current = 12 mA, room temperature, 3 h, isolated yields''

    \vspace{3mm}
    
    \textbf{Answer:} ''Bu4NBF4''
\end{tcolorbox}

\paragraph{Synthetic Reaction Solvent Extraction(SolvE):} In this task, models are challenged to accurately identify and report the solvent used in chemical synthesis processes from given texts. Solvents play a critical role in executing and regulating chemical reactions, making their extraction essential for detailed analysis.

\begin{tcolorbox}[title=Prompt Example]

    \textbf{Query:} You are an expert in the field of chemistry and here is a task to extract information from the chemistry literature that requires you to extract the Solvent (Op) involved from the input.\\
    You must output your prediction, i.e. an effective chemical agent, and follow the output format exactly as follows:\\
    \{''answer'': ''an effective chemical agent ''\}. I don't need any explanation, you just need to output your judgment in format.\\
    The contents of input are as follows:\\
    ''input'':''For instance, acetone served as a nucleophile to react with 2H-azirines under the basic conditions to furnish pyrroles. Upon changing the catalyst to TfOH, acetone served as an electrophile to synthesize 3-oxazolines. Moreover, the products could be synthesized on a gram scale, and the possible catalytic cycles were proposed.''

    \vspace{3mm}
    
    \textbf{Answer:} ''Acetone''
\end{tcolorbox}

\paragraph{Reaction Temperature Extraction(TempE):} In this task, models are required to accurately identify and report the specific temperature in Celsius at which a chemical synthesis takes place, as detailed within the textual descriptions. This parameter is crucial for defining reaction conditions.

\begin{tcolorbox}[title=Prompt Example]

    \textbf{Query:} You are an expert in the field of chemistry and here is a task to extract information from the chemistry literature that requires you to extract the Temperature (Op) involved from the input.\\
    You must output your prediction, i.e. a specific Celsius temperature, and follow the output format exactly as follows:\\
    \{''answer'': ''a specific Celsius temperature ''\}. I don't need any explanation, you just need to output your judgment in format.\\
    The contents of input are as follows:\\
    ''input'': ''The reaction was carried out with 1a (0.10mmol, 1.0 equiv), 2 (1.0 mL, 136 equiv), Zn(OTf)2 (10 mol \%), and NaOH (120 mol \%) at 56 $\tccentigrade$ for 48 h.''

    \vspace{3mm}
    
    \textbf{Answer:} ''56 $\tccentigrade$''
\end{tcolorbox}

\paragraph{Reaction Time Extraction(TimeE):} In this task, models must precisely extract and report the duration of chemical synthesis reactions in hours, directly from the literature, adhering to a specified output format.

\begin{tcolorbox}[title=Prompt Example]

    \textbf{Query:} You are an expert in the field of chemistry and here is a task to extract information from the chemistry literature that requires you to extract the Reaction time (Op) involved from the input.\\
    You must output your prediction, i.e. a specific time (in h), and follow the output format exactly as follows:\\
    \{''answer'': ''a specific time (in h) ''\}. I don't need any explanation, you just need to output your judgment in format.\\
    The contents of input are as follows:\\
    ''input'': ''The reaction was carried out with 1a (0.10mmol, 1.0 equiv), 2 (1.0 mL, 136 equiv), Zn(OTf)2 (10 mol \%), and NaOH (120 mol \%) at 56 $\tccentigrade$ for 48 h.''

    \vspace{3mm}
    
    \textbf{Answer:} ''48 h''
\end{tcolorbox}

\paragraph{Reaction Product Extraction(ProdE):} This task demands models to adeptly identify and categorize chemical product entities within descriptions of synthesis reactions. Utilizing Named Entity Recognition (NER) techniques, models must accurately recognize product terms, reflecting the end results of chemical processes, in scientific literature analysis.

\begin{tcolorbox}[title=Prompt Example]

    \textbf{Query:} You are an expert in the field of chemistry and will be required to perform named entity recognition (ner) task. The following paragraph is a string list of a sentence of a paper from which you need to combine the information in the sentence to extract the important entities and give them entity labels. For each element of the sentence list, an entity label is required. \\
    The allowed entity labels are B-Prod, I-Prod, O. Note: B-Prod indicates the beginning character of the chemical reaction product entity, I indicates the internal character of the chemical reaction product entity, and O indicates that it does not belong to any entity. \\
    You need to determine the type of each element in the sentence list according to the entity type above and fill the corresponding entity label in the output list, even if the element in the list is not a word, you need to determine and output. You must output your prediction, i.e. a list , and follow the output format exactly as follows:\{''answer'': ''a list ''\}. Next is the sentence list: \\
    \text{[`;', `Rauchfuss', `,', `T.']} Please answer in the following format strictly: [a, b, c... ](where a, b, c represent possible labels); Do not output any explanatory text, only output the list. please ensure the output list length matches the input list length.

    \vspace{3mm}
    
    \textbf{Answer:} ''[`O', `O', `O', `O']''
\end{tcolorbox}

\paragraph{Characterization Method Extraction(CharME):} In this task, models are required to identify and summarize the analytical techniques, including spectroscopy and chromatography, utilized for characterizing chemical compounds within research texts.

\begin{tcolorbox}[title=Prompt Example]

    \textbf{Query:} You are an expert in the field of chemistry. This task involves extracting information from chemical literature and categorizing the characterization methods of the related reactions. The classification methods you can choose from include: 1. Nuclear Magnetic Resonance (NMR) Spectroscopy; 2. Mass Spectrometry (MS); 3. Infrared Spectroscopy (IR); 4. Ultraviolet-Visible Spectroscopy (UV-Vis); 5. X-Ray Diffraction (XRD); 6. Gas Chromatography (GC) and High Performance Liquid Chromatography (HPLC); 7. Elemental Analysis (EA); 8. Thermal Analysis; 9. Electrochemical Analysis; 10. X-Ray Photoelectron Spectroscopy (XPS/ESCA); 11. Raman Spectroscopy; 12. Transmission Electron Microscopy (TEM) and 13. Scanning Electron Microscopy (SEM), etc. You must output in the following format: \{''answer'': ''Characterization Methods''\}. The characterization methods can be one or more of the above. I don't need any explanations, you just need to output the answer according to the format. Here is the input content:''input'': ''H NMR, 13C NMR and 19F NMR spectra were recorded on Bruker 400 MHz, 500 MHz or 600 M spectrometer at 295 K in CDCl3 unless otherwise noted. Data for 1H NMR were reported as follows: chemical shift ($\delta$
 ppm), multiplicity, coupling constant (Hz), and integration. Data for 13C NMR were reported as follows: chemical shift ($\delta$
 ppm), multiplicity, and coupling constant (Hz). Data for 19F NMR were reported as follows: chemical shift ($\delta$
 ppm), multiplicity, coupling constant (Hz). Chemical shifts were reported using the residual solvent CHCl3 as the internal reference for 1H NMR ($\delta$
 = 7.260 ppm) and CDCl3 peak as the internal reference for 13C NMR ($\delta$
 = 77.160 ppm). High-resolution mass spectral analysis (HRMS) data were acquired on Water XEVO G2 Q-TOF (Waters Corporation) or Gas Chromatography Q Exactive Orbitrap Mass Sepctrometer (Thermo-Fisher). Gas chromatographic (GC) analysis was acquired on a Shimadzu GC-2010 plus Series GC system equipped with a flame-ionization detector. Organic solutions were concentrated under reduced pressure on Buchi rotary evaporator. Column chromatographic purification of products was accomplished using forced-flow chromatography on Silica Gel (300-400 mesh).''

    \vspace{3mm}
    
    \textbf{Answer:} ''Nuclear Magnetic Resonance (NMR) Spectroscopy,Mass Spectrometry (MS),Electrochemical Analysis''
\end{tcolorbox}

\paragraph{Catalysis Type Extraction(CatTE):} The objective of this task is to identify and classify the type of catalysis—homogeneous or heterogeneous—utilized in chemical reactions as depicted in literature. Models are tasked with accurately determining the catalyst category and reporting it in a consistent format.

\begin{tcolorbox}[title=Prompt Example]

    \textbf{Query:} You are an expert in the field of chemistry and here is a task to extract information from the chemistry literature that requires you to extract the Catalytic method(Reaction driving force) involved from the input.\\
    You must output your prediction, i.e. a specific Catalytic method (such as Thermocatalytic reaction, Electrocatalytic reaction etc.), and follow the output format exactly as follows:\\
    \{''answer'': ''a specific Catalytic method ''\}. I don't need any explanation, you just need to output your judgment in format.\\
    The contents of input are as follows:\\
    ''input'': ''The reaction was carried out with 1a (0.10 mmol, 1.0 equiv), 2 (1.0 mL, 136 equiv), Zn(OTf)2 (10 mol \%), and NaOH (120 mol \%) at 56 $\tccentigrade$ for 48 h''

    \vspace{3mm}
    
    \textbf{Answer:} ''Thermocatalytic reaction''
\end{tcolorbox}

\paragraph{Yield Extraction(YieldE):} This task centers on identifying and quantifying the yield of chemical syntheses—the efficiency metric indicating the amount of desired product produced. Models must accurately extract and report yield percentages from chemistry literature in a numerical format.

\begin{tcolorbox}[title=Prompt Example]

    \textbf{Query:} You are an expert in the field of chemistry and here is a task to extract information from the chemistry literature that requires you to extract the Reaction Yield involved from the input.\\
    You must output your prediction, i.e. a specific percentage, and follow the output format exactly as follows:\\
    \{''answer'': ''a specific percentage ''\}. I don't need any explanation, you just need to output your judgment in format.\\
    The contents of input are as follows:\\
    ''input'': ''Finally, pyrrole 3a was isolated in an 81\% yield by increasing the loading of NaOH to 1.2 equiv and prolonging the reaction time to 48 h (entry 17).''

    \vspace{3mm}
    
    \textbf{Answer:} ''up to 81\%''
\end{tcolorbox}

\subsubsection{Inductive Generation}
\paragraph{Chemical Paper Abstract Generation(AbsGen):} This task entails automating the creation of succinct abstracts for chemistry research papers, effectively summarizing core methodologies, findings, and implications in a brief format.

\begin{tcolorbox}[title=Prompt Example]

    \textbf{Query:} You are an expert in the field of chemistry and here is an assignment for you to summarise the chemistry literature and generate information. The requirement is to summarise the abstract (including methods, conclusions and implications) of the input paper in a short paragraph, which can be summarised as: someone did something, used some methods, what were the conclusions, and what were the implications. You must output your answer in the following format, that is, specific post-processing method: \{`answer': `the abstract which you summarize'\}. I don't need anything unrelated to this, you just need to output as required. The input is given below:\\

    \textit{The complete information of a paper, including title, authors, citation, access, abstract, main body, and references. The actual information from the paper ''Palladium-Catalyzed Divergent Enantioselective Functionalization of Cyclobutenes” is used here.}

    \vspace{3mm}
    
    \textbf{Answer:} ''Ping Lu's group has realized three asymmetric functionalization reactions of cyclobutene, including hydroarylation, 1,3-bisarylation, and 1,2-bisarylation, using transition metal palladium/BI-DIME-type phosphine chiral ligands. The enantioselectivity and regioselectivity of the reactions were elucidated by controlled experiments and DFT calculations, and the mechanism of selectivity at the deuterium substitution site was explained, realising the preparation of chiral cyclobutane derivatives with multiple substitution modes via the same intermediate.''
\end{tcolorbox}

\paragraph{Research Outline Generation(OLGen):} This task demands the creation of structured outlines for chemical research papers, systematically organizing key points and findings in a coherent sequence that mirrors the document's content. The model must distill information from literature and arrange it logically to reflect the research narrative.

\begin{tcolorbox}[title=Prompt Example]

    \textbf{Query:} You are an expert in the field of chemistry. This is a task to extract information from chemical literature, requiring you to extract the outline of the research content of the paper from the input, specifically what each part of this section did, and output it in the order of 1, 2, 3, 4. You must output your answer in the following format, that is, specific post-processing method: {'answer': '1. Outline 1; 2. Outline 2; 3. Outline 3, and so on'}. I don't need any explanations, just output the answer according to the format. The input content is as follows:\\

    \textit{The complete information of a paper, including title, authors, citation, access, abstract, main body, and references. The actual information from the paper ''Borrowing Hydrogen $\beta$
-Phosphinomethylation of Alcohols Using Methanol as C1 Source by Pincer Manganese Complex” is used here.}

    \vspace{3mm}
    
    \textbf{Answer:} ''1. Reaction substrate range; 2. Mechanism experiments; 3. Cyclic voltammetry experiments; 4. Gram-scale preparation.''
\end{tcolorbox}

\paragraph{Chemical Literature Topic Classification(TopC):} This task involv-es expertly categorizing chemical literature into specific research areas, facilitating organized information retrieval by subject. It requires discerning the primary focus of content and assigning it a fitting label from a predefined list of chemical disciplines.

\begin{tcolorbox}[title=Prompt Example]

    \textbf{Query:} You are an expert in the field of Chemistry and here is an assignment that requires you to classify chemistry literature. Your optional classification types are Chemistry Education, Organic Chemistry, Materials, Nanotechnology, Natural Products, Polymer Chemistry, Medicinal Chemistry, Life Sciences, Physical Chemistry, Theoretical and Computational Chemistry, Inorganic Chemistry, Environmental Science, Analytical Chemistry, Food and Daily Products, Industry and Commerce. Please select the type of literature from these types that best matches the input content. \\
    You must output your prediction, i.e. a specific type, and follow the output format exactly as follows:\\
    \{''answer'': ''a specific type ''\}. I don't need any explanation, you just need to output your judgment in format.\\
    The input is as follows: Phosphagermylenylidenes (R-P=Ge), as heavier analogs of isonitriles, whether in their free state or as complexes with a Lewis base, have not been previously identified as isolable entities. In this study, we report the synthesis of a stable monomeric phosphagermylenylidene within the coordination sphere of a Lewis base under ambient conditions. This species was synthesized by Lewis base-induced dedimerization of a cyclic phosphagermylenylidene dimer or via Me3SiCl elimination from a phosphinochlorogermylene framework. The deliberate integration of a bulky, electropositive N-heterocyclic boryl group at the phosphorus site, combined with coordination stabilization by a cyclic (alkyl)(amino)carbene at the low-valent germanium site, effectively mitigated its natural tendency toward oligomerization. Structural analyses and theoretical calculations have demonstrated that this unprecedented species features a P=Ge double bond, characterized by conventional electron-sharing $\pi$ and $\sigma$ bonds, complemented by lone pairs at both the phosphorus and germanium atoms. Preliminary reactivity studies show that this base-stabilized phosphagermylenylidene demonstrates facile release of ligands at the Ge atom, coordination to silver through the lone pair on P, and versatile reactivity including both (cyclo)addition and cleavage of the P=Ge double bond.

    \vspace{3mm}
    
    \textbf{Answer:} ''Organic Chemistry''
\end{tcolorbox}

\paragraph{Reaction Type Recognition and Induction(ReactTR):} This task centers on identifying diverse chemical reaction types from literature and deducing overarching principles or patterns. It requires recognizing the specific reaction type in given excerpts and succinctly reporting it in a standardized format.

\begin{tcolorbox}[title=Prompt Example]

    \textbf{Query:} You are an expert in the field of chemistry and here is a task to extract information from the chemistry literature that requires you to extract the Reaction type involved from the input.\\
    You must output your prediction, i.e. a specific Reaction type, and follow the output format exactly as follows:\\
    \{''answer'': ''a specific Reaction type ''\}. I don't need any explanation, you just need to output your judgment in format.\\
    The contents of input are as follows:\\
    ''input'': ''An electrochemical oxidative alkoxysulfonylation of alkenes using sulfonyl hydrazines with alcohols is accomplished, which leads to the easy accessibility of $\beta$
-alkoxy sulfones, which are not only valuable architectures of many biologically active molecules, but also key building blocks for various organic transformations.''

    \vspace{3mm}
    
    \textbf{Answer:} ''electrochemical oxidative''
\end{tcolorbox}

\subsection{Molecular Understanding}

\subsubsection{Molecular Name Generation}
\paragraph{Molecular Name Generation from Text Description(MolNG):} This task assesses the ability of LLMs to generate valid chemical structure representations, specifically SMILES strings, from complex textual descriptions of molecular structures, properties, and classifications. It requires converting such detailed text inputs into accurate names of chemical molecules.

\begin{tcolorbox}[title=Prompt Example]

    \textbf{Query:} You are an expert in the field of chemistry and now have a ''molecular generation'' task. The main content of this task is to generate valid SMILES information that matches the description of the molecule provided.\\
    Specific mission questions are as follows:\\
    I need a molecule that meets the following conditions: The molecule is an imidazole that is 1H-imidazole substituted at position 2 by a nitro group. It has a role as an antitubercular agent. It is a C-nitro compound and a member of imidazoles. It derives from a 1H-imidazole. Please represent the molecule in SMILES.\\
    You must output your prediction, i.e. a valid SMILES, and follow the output format exactly as follows:\\
    \{''answer'': ''a valid SMILES ''\}. I don't need any explanation, you just need to output your judgment in format.

    \vspace{3mm}
    
    \textbf{Answer:} ''O=[N+]([O-])C1=NC=CN1''
\end{tcolorbox}

\subsubsection{Molecular Name Translation}
\paragraph{Molecular Representation Name Translation:} This task encompas-ses five sub-tasks, each requiring models to translate between molecular naming conventions, including conversions from IUPAC to Molecular Formula, SMILES to Molecular Formula, IUPAC to SMI-LES, SMILES to IUPAC, and the bidirectional translation between SMIL-ES and SELFIES. Given the uniformity of sub-task prompts except for the specific conversion directives, this section exemplifies the process using the IUPAC to Molecular Formula transformation.

\begin{tcolorbox}[title=Prompt Example]

    \textbf{Query:} Please tell me the molecular formula of the following molecules given by IUPAC. The molecule is N-[3-[3-(4-phenoxyphenyl)-1,2,4-oxadiazol-5-yl]phenyl]acetamide.You must output your prediction, i.e. a valid molecular formula, and follow the output format exactly as follows:\\
    \{''answer'': ''The answer you judge ''\}. I don't need any explanation, you just need to output your judgment in format.

    \vspace{3mm}
    
    \textbf{Answer:} ''C22H17N3O3''
\end{tcolorbox}

\subsubsection{Molecular Property Prediction}
\paragraph{Molecule Property classification (MolPC):} This task aims to predict molecular property categories based on a provided SMILES representation. Properties include ClinTox, HIV inhibition, Blood-Brain Barrier (BBBP) penetration, Side Effect Resource (SIDER) effects, and polarity. An example prompt for predicting ClinTox classification is illustrated below.

\begin{tcolorbox}[title=Prompt Example]

    \textbf{Query:} You are an expert in the field of chemistry and now have a ClinTox prediction task. \\
    The ClinTox (Clinical Toxicity) prediction task aims to predict the toxicity problems that compounds may cause in clinical trials, thereby avoiding potential drug development failures and ensuring patient safety. Specific mission questions are as follows:\\
    c1cc(cc(c1)[131I])C[NH+]=C(N)N\\
    You have to output your guess, which is ''Yes'' or ''No'', and exactly follow the output format, which is as follows:\\
    \{''answer'': ''The answer you judge''\}. I don't need any explanation, you just need to output your judgment in the format.

    \vspace{3mm}
    
    \textbf{Answer:} ''Yes''
\end{tcolorbox}

\paragraph{Molecule Property Regression (MolPR):} This task focuses on forecasting numerical values related to molecular properties, encompassing HOMO and LUMO energies, ESOL, lipophilicity, polarity, melting and boiling points. The model is tasked with predicting these values based on the representation of input molecules. As an illustration, we present a prompt for predicting HOMO energy levels.

\begin{tcolorbox}[title=Prompt Example]

    \textbf{Query:} You are an expert in the field of chemistry. I have a molecular SELFIES, which is shown below. \\{}
    [O][C][C][O][C][Branch1][Ring1][C][\#C][C][Ring1]\\{}
    [\#Branch1][O]\\
    Please provide me with the HOMO energy value of this molecule.\\
    You do not need to output too much explanation, you must output a number, and follow the following output format:\\
    \{''answer'': ''a specific number ''\}

    \vspace{3mm}
    
    \textbf{Answer:} ''-0.2631''
\end{tcolorbox}

\subsubsection{Molecular Description}
\paragraph{Physicochemical Property Prediction from Molecular Structure (Mol2PC):} This task demands the model to predict the physicochemical properties of molecules solely from their SMILES or SELFIES representations, leveraging chemical insights to provide detailed descriptions.

\begin{tcolorbox}[title=Prompt Example]

    \textbf{Query:} You are an expert chemist. Given the molecule SMILES, your task is to provide the detailed description of the molecule using your experienced chemical knowledge.\\
    Please strictly follow the format, no other information can be provided.\\
    The Molecule SMILES is as follows:\\
    C1=CC=C2C(=C1)NC1=CN=CC=C12 You must output your prediction, i.e. a description of the molecule, and follow the output format exactly as follows:\\
    \{''answer'': ''The answer you judge ''\}. I don't need any explanation, you just need to output your judgment in format.

    \vspace{3mm}
    
    \textbf{Answer:} ''The molecule is the parent compound of the beta-carbolines, a tricyclic structure comprising an indole ring system ortho- fused to C-3 and C-4 of a pyridine ring. It has a role as a marine metabolite and a fungal metabolite. It is a member of beta-carbolines and a mancude organic heterotricyclic parent.''
\end{tcolorbox}

\subsection{Scientific Knowledge Deduction}

\subsubsection{Retrosynthesis Analysis}
\paragraph{Substrate Recommendation (SubRec):} This task challenges the m-odel to complete an incomplete chemical equation by recommending suitable reactants, requiring inference of the missing components in the given expression.

\begin{tcolorbox}[title=Prompt Example]

    \textbf{Query:} Chemical reaction equations are typically expressed in the following form: reactant1.reactant2.reactant3...>>product. In this form, each substance(reactant/product) is represented using the SMILES notation. Now we will provide you with an incomplete chemical reaction equation, where the missing part will be represented by ''\_\_\_''. The missing parts could be one or more substances. Based on the remaining portions of the reaction equation, please infer what the missing part could be. Please only provide the missing part in your response, without any additional content.\\
    The Incomplete equation is as follows:\\
    \_\_\_>>OC(C1=CC=CC=C1)(C1=CC=NC=C1)C1=CC=CC=\\
    C1C1=CC=CC=C1\\
    You must output your prediction, i.e. valid SMILES, and follow the output format exactly as follows:\\
    \{''answer'': ['' valid SMILES 1 '', '' valid SMILES 2 '', ...]\}. \\
    I don't need any explanation, you just need to output your judgment in format. 

    \vspace{3mm}
    
    \textbf{Answer:} ''[`BrC1=CC=C(C2=CC=CC=C2)C=C1',`O=C(C1=\\
    CC=CC=C1)C1=CC=NC=C1']''
\end{tcolorbox}

\paragraph{ Synthetic Pathway Recommendation (PathRec):} This task challenges the model to propose synthesis pathways, ranging from single-step to multi-step processes, connecting given reactants to products. As an illustration, the prompt for recommending a multi-step synthesis path is highlighted.

\begin{tcolorbox}[title=Prompt Example]

    \textbf{Query:} You are an expert in the field of chemistry and now have a multi-step synthesis route recommendation task. \\
    If starting from methylbenzyl ketone as the raw material to synthesize the product tacrine, can you provide a multi-step synthesis route?\\
    You have to output your prediction, and exactly follow the output format, which is as follows:\\
    \{''answer'': ''1.The first step you judge,2.The second step you judge,...''\}.  I don't need any explanation, you just need to output your judgment in the format.

    \vspace{3mm}
    
    \textbf{Answer:} ''1. Methylbenzyl ketone reacts with aminoacetaldehyde under acidic conditions to produce tetrahydroisoquinoline. \\
    2. Tetrahydroisoquinoline undergoes a Pfitzinger reaction with maleic acid under acidic conditions to yield tacrine.''
\end{tcolorbox}

\paragraph{Synthetic Difficulty Evaluation(SynDE):} This task centers on evaluating the synthetic challenge of given compounds, using their Synthetic Accessibility Scores (SAS) as the basis for assessment.

\begin{tcolorbox}[title=Prompt Example]

    \textbf{Query:} You are an expert in the field of chemistry and now have a difficult evaluation task. \\
    Based on the Synthetic Accessibility score, evaluate how difficult it is to synthesize the compound, which has a SMILES of: \\{}
    [H]C([H])([H])C([H])([H])C([H])([H])C([H])([H])C([H])\\
    ([H])C([H])([H])C([H])([H])C([H])([H])C([H])([H])C([H])\\
    ([H])C([H])([H])C([H])([H])C([H])([H])C([H])([H])OS\\
    ([O])([O])C(F)(F)F\\
    You have to output your prediction, and exactly follow the output format, which is as follows:\\
    \{''answer'': ''The answer you judge''\}.  I don't need any explanation, you just need to output your judgment in the format.

    \vspace{3mm}
    
    \textbf{Answer:} ''2.7''
\end{tcolorbox}

\subsubsection{Reaction Condition Recommendation}
\paragraph{Ligand, Reagent, and Solvent Recommendation(LRec, RRec, SolvR-ec):} The task is given an incomplete chemical reaction represented by SMILES, and it requires the large model to fill in the blank in the form of a multiple-choice question. The prompts for the three recommendation sub-tasks are completely similar in form, and here we use ligand recommendation as an example.

\begin{tcolorbox}[title=Prompt Example]

    \textbf{Query:} You are an expert chemist. Here is an incomplete chemical reaction: \\
    Clc1ccc2ncccc2c1.Cc1ccc2c(cnn2C2CCCCO2)c1[B-](F)\\(F)F.[K+] >\_\_\_.CO.CCN(CC)
    CC>Unknown product.\\
    Please fill in the blank (''\_\_\_'') using the optimal ligand from the CHOICE LIST to maximize the yield of the reaction.\\
    CHOICE LIST: (Choose from below!)\\
    > c1ccc(P(c2ccccc2)c2ccccc2)cc1\\
    > CN(C)Cc1ccc(P(C(C)(C)C)C(C)(C)C)cc1\\
    > C1CCC(P(C2CCCCC2)C2CCCCC2)CC1\\
    > Cc1ccccc1P(c1ccccc1C)c1ccccc1C\\
    > CCCCP(C12CC3CC(CC(C3)C1)C2)C12CC3CC(CC(C3)C1)\\C2\\
    > COc1cccc(OC)c1-c1ccccc1P(C1CCCCC1)C1CCCCC1\\
    > CC(C)(C)P([C]1C=C[CH][C@@H]1[Fe]C1C=CC=C1P(C\\(C)(C)C)C(C)(C)C)C(C)(C)C\\
    > CC(C)c1cc(C(C)C)c(-c2ccccc2P(C2CCCCC2)C2CCCCC2)\\c(C(C)C)c1\\
    > C1=C[C@H]2[Fe][C@@H]1[C]2P(c1ccccc1)c1ccccc1\\
    > CC1(C)c2cccc(P(c3ccccc3)c3ccccc3)c2Oc2c(P(c3ccccc3)c3\\ccccc3)cccc21\\
    Answer: (Choose from the CHOICE LIST!)\\
    You must output your prediction, i.e. valid SMILES, and follow the output format exactly as follows:\\
    \{''answer'': ['' valid SMILES'']\}. \\
    I don't need any explanation, you just need to output your judgment in format.

    \vspace{3mm}
    
    \textbf{Answer:} ''c1ccc(P(c2ccccc2)c2ccccc2)cc1''
\end{tcolorbox}

\paragraph{Catalyst, Reaction Temperature, and Reaction Time Recommendation(CatRec, TempRec, TimeRec):} The task requires the model to recommend the catalyst, reaction temperature, or reaction time for a reaction based on the provided reactants, products, and reaction type (e.g., Buchwald coupling reaction). The prompts for the three recommendation sub-tasks are completely similar in form, and here we use the prompt for catalyst recommendation as an example.

\begin{tcolorbox}[title=Prompt Example]

    \textbf{Query:} You are an expert in the field of chemistry, and there will be a catalyst recommendation task below, where you will recommend the catalyst you feel is suitable based on the reaction information provided. The questions are as follows:\\
    For the buchwald coupling reaction, the reactants are O=C1NCc2c(OCCCCN3CCNCC3)cccc21 and Brc1ccc2ncccc2c1, and the product is O=C1NCc2c(OCCCCN3CCN(CC3)c4ccc5ncccc5c4)cccc12. Give recommendations for catalyst.\\
    You should strictly follow the output requirements, the output format is as follows: \{`answer': ` effective catalyst '\}. You simply output as required, no additional explanatory information is required.

    \vspace{3mm}
    
    \textbf{Answer:} ''C(O[Pd]OC(C)=O)(C)=O and c1(c2c(P(C3CCCC\\
    C3)C3CCCCC3)cccc2)c(cc(cc1C(C)C)C(C)C)C(C)C''
\end{tcolorbox}

\subsubsection{Reaction Outcome Prediction}

\paragraph{Reaction Product Prediction (PPred):} This task involves predicting the product SMILES of a chemical reaction, given the reactant SMILES and specified reaction conditions.

\begin{tcolorbox}[title=Prompt Example]

    \textbf{Query:} You are an expert in the field of chemistry and now have a ''forward synthesis'' task. The main content of the task is to predict SMILES, the product of a chemical reaction, based on the provided reactant SMILES (Use ''.'' To divide the SMILES.).\\
    Specific mission questions are as follows:\\
    C(=NC1CCCCC1)=NC1CCCCC1.CC\#N.CC(=O)OC[C@H]\\
    (NC(=O)[C@H](CC(C)C)NC(=O)OCC1=CC=CC=C1)[C@\\
    @H](OC(C)=O)[C@@H](OC(C)=O)[C@H](OC(C)=O)\\
    C(=O)O.CCN(CC)CC.CCOC(=O)C[C@H](N)C1=CC=C(C)\\C=C1 
    Given the above reactants and reagents, what could be a probable product of their reaction?\\
    You must output your prediction, i.e. valid SMILES, and follow the output format exactly as follows:\\
    \{''answer'': '' valid SMILES ''\}. (Use ''.'' To divide the SMILES.) I don't need any explanation, you just need to output your judgment in format.

    \vspace{3mm}
    
    \textbf{Answer:} ''CCOC(=O)C[C@H](NC(=O)[C@@H](OC(C)=O)\\{}
    [C@H(OC(C)=O)[C@H](OC(C)=O)[C@H](COC(C)=O)NC\\
    (=O)[C@H](CC(C)C)NC(=O)OCC1=CC=CC=C1)C1=CC=C\\(C)C=C1''
\end{tcolorbox}

\paragraph{Product Yield Prediction (YPred):} This task centers on predicting whether a chemical reaction will yield high or non-high yields, based on the supplied reaction SMILES string. This forecast aids in optimizing the synthesis process by anticipating reaction efficiency under given conditions.

\begin{tcolorbox}[title=Prompt Example]

    \textbf{Query:} You are an expert chemist, your task is to predict the yield of reaction using your experienced chemical yield prediction knowledge.\\
    Please strictly follow the format, no other information can be provided. Given the SMILES string of a Suzuki reaction, the task focuses on predicting reaction yield, specifically whether a reaction is High-yielding (Yes) or Not High-yielding (No), based on the SMILES string representation of each Suzuki reaction. The reactants are separated by `.', which are separated from the product by `>>'. High-yielding reaction means the yield rate of the reaction is above 70. You will be provided with several examples molecules. Please answer with only Yes or No. The Reaction is as follows\\
    C1CCOC1.CC(=O)O~CC(=O)O~[Pd].CCc1cccc(CC)c1.CN(C)\\
    c1ccc(P(C(C)(C)C)C(C)(C)C)cc1.Cc1ccc2c(cnn2C2CCCC\\
    O2)c1[B-](F)(F)F.O.O=S(=O)(Oc1ccc2ncccc2c1)C(F)(F)F.\\
    \text{[}K+]>>Cc1ccc2c(cnn2C2CCCCO2)c1-c1ccc2ncccc2c1 You have to output your judge, which is ''Yes'' or ''No'', and exactly follow the output format, which is as follows:\\
    \{''answer'': ''The answer you judge''\}. I don't need any explanation, you just need to output your judgment in the format.

    \vspace{3mm}
    
    \textbf{Answer:} ''No''
\end{tcolorbox}

\paragraph{Reaction Rate Prediction (RatePred):} This task focuses on predicting the approximate range of activation energy for a reaction under specified conditions, essential for comprehending reaction kinetics and optimizing operational parameters.

\begin{tcolorbox}[title=Prompt Example]

    \textbf{Query(Translation):}
    You are an expert in the field of chemistry. Below is a task about characterizing the activation energy of a reaction. Your task is to predict the approximate range of the reaction activation energy based on the provided information. The question is as follows: Under standard temperature and pressure (298 K, 1 atm), what is the approximate range of the activation energy for the dehydration of ethanol to form ethylene? Please answer strictly in the following format: \{''answer'': ''your predicted range''\}. You only need to output your prediction without any additional explanations or introductions.

    \vspace{3mm}
    
    \textbf{Answer:} ''180-370 kJ/mol''
\end{tcolorbox}

\subsubsection{Reaction Mechanism Analysis}
\paragraph{Intermediate Derivation(IMDer):} This task involves the large language model deriving potential intermediates that form during a chemical reaction, based on provided reactants and products. These intermediates are transient species, relatively stable compared to transition states, yet distinct from the final products.

\begin{tcolorbox}[title=Prompt Example]

    \textbf{Query:} You are an expert in the field of chemistry and there is a task to derive reaction intermediates below. Intermediates are molecules or ions formed during chemical reactions that are stable but not the final product. In the hydrolysis of tert-butyl chloride, what type of intermediate forms? \\
    You should output exactly as follows, \{'' answer '':'' Your prediction ''\}. You only need to output the requested content, no additional explanation content. 

    \vspace{3mm}
    
    \textbf{Answer:} ''Carbocation''
\end{tcolorbox}


\end{document}